\documentclass[runningheads]{llncs}
\usepackage{graphicx}
\usepackage{comment}
\usepackage{amsmath,amssymb} 
\usepackage{color}

\usepackage{epsfig}
\usepackage{textcomp}
\usepackage{subcaption}
\usepackage{tabulary}
\usepackage{epigraph}
\usepackage{xspace}
\usepackage{multirow}
\usepackage[table]{xcolor}
\definecolor{grey}{rgb}{0.9,0.9,0.9}
\usepackage{ctable}
\usepackage{algorithm}
\usepackage{algorithmic}
\usepackage{wrapfig}

\makeatletter
\DeclareRobustCommand\onedot{\futurelet\@let@token\@onedot}
\def\@onedot{\ifx\@let@token.\else.\null\fi\xspace}

\def\eg{\emph{e.g}\onedot} 
\def\ie{\emph{i.e}\onedot}

\makeatother

\begin{document}
\pagestyle{headings}
\mainmatter
\def\ECCVSubNumber{3672}  

\title{DSDNet: Deep Structured self-Driving Network} 

\titlerunning{DSDNet: Deep Structured self-Driving Network}
%
\author{Wenyuan Zeng\inst{1,2} \and
Shenlong Wang\inst{1,2} \and
Renjie Liao\inst{1,2} \and \\
Yun Chen\inst{1} \and
Bin Yang\inst{1,2}\and
Raquel Urtasun\inst{1,2}
}
\authorrunning{W. Zeng et al.}
\institute{Uber ATG \and University of Toronto\\
\email{\{wenyuan,slwang,rjliao,yun.chen,byang10,urtasun\}@uber.com}}
\maketitle

\begin{abstract}

In this paper, we propose the Deep Structured self-Driving Network (DSDNet), which performs object detection, motion prediction, and motion planning with a single neural network. 
Towards this goal, we develop a deep structured energy based model which considers the interactions between actors and produces
socially consistent multimodal future predictions.
Furthermore, DSDNet explicitly exploits the predicted future distributions of actors to plan a safe maneuver by using a structured planning cost.
Our sample-based formulation allows us to overcome the difficulty in probabilistic inference of continuous random variables. 
Experiments on a number of large-scale self driving datasets demonstrate that our model significantly outperforms the state-of-the-art.

\keywords{Autonomous driving, motion prediction, motion planning}

\end{abstract}


\section{Introduction}


The self-driving problem can be described as  safely, comfortably and efficiently maneuvering a vehicle from point A to point B.
This task is very complex; Even the most intelligent agents to date (\ie, humans) are very frequently involved in traffic accidents. Despite
the development of Advanced Driver-Assistance Systems (ADAS), 1.3 million people die every year on the road, and 20 to 50 million are severely injured. 

Avoiding collisions in complicated traffic scenarios is not easy, primarily due to the fact that there are other traffic participants, whose future behaviors are unknown and very hard to predict. 
A vehicle that is next to our lane and blocked by its leading vehicle might decide to stay in its lane or cut in front of us.
A pedestrian waiting on the edge of the road might decide to cross the road at any time. 
Moreover, the behavior of each actor depends on the actions taken by other actors, making
the prediction task even harder.
Thus,  it is extremely important to model the future motions of actors with multi-modal  distributions 
that also consider the interactions between actors.

To safely drive on the road, a self-driving vehicle (SDV) needs to detect surrounding actors, predict their future behaviors, and plan safe maneuvers.
Despite the recent success of  deep learning for perception, the prediction task, due to the aforementioned challenges, remains an open problem.
Furthermore, there is also a need to develop motion planners that can take the uncertainty of the predictions  into account. 
Previous works have utilized parametric distributions to model multimodality of motion prediction.   Mixture of Gaussians
\cite{chai2019multipath,hong2019rules} are a natural approach due to their close-form 
 inference. 
 However, it is hard to decide the number of modes in advance. Furthermore, these approaches suffer from mode collapse during training
 \cite{jain2019discrete,hong2019rules,rhinehart2018r2p2}. 
An alternative is to learn a model distribution from data using, \eg, neural networks. As shown in \cite{rhinehart2019precog,lee2017desire,tang2019multiple}, a CVAE \cite{sohn2015learning} can be applied to
capture multi-modality, and the interactions between actors can be modeled through latent variables.
However, it is typically hard/slow to do probabilistic inference and the interaction mechanism does not explicitly model
collision which humans want to avoid at all causes. Besides, none of these works have shown the effects upon planning
on real-world datasets.

\begin{figure}[t]
\begin{center}
  \includegraphics[height=3.0cm]{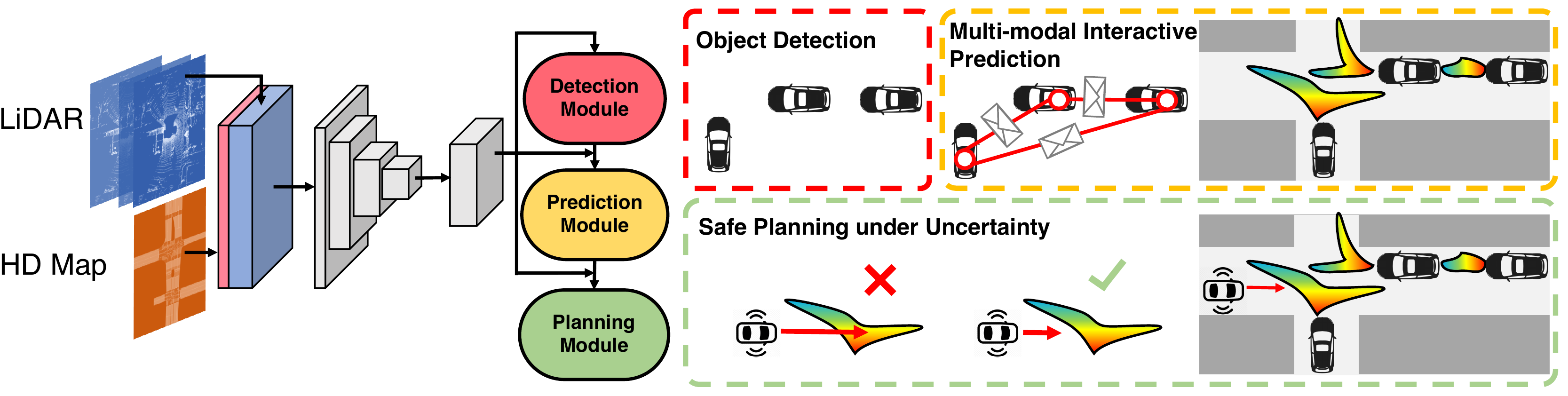}
\end{center}
\caption{\textbf{DSDNet overview:} The model takes LiDAR and map as inputs, processes them with a CNN backbone, and jointly performs object detection,
multi-modal socially-consistent prediction, and safe planning under uncertainty. \textit{Rainbow} patterns mean highly likely actors' future positions
predicted by our model.}
\label{fig:model}
\end{figure}

In this paper we propose the \textbf{D}eep \textbf{S}tructured self-\textbf{D}riving \textbf{Net}work (DSDNet), 
a single neural network that takes raw sensor data as input to jointly detect actors in the scene, predict a multimodal distribution over their
future behaviors, and produce safe plans for the SDV. 
This paper has three key contributions: 
\begin{itemize}
  \item Our prediction module uses an energy-based formulation to explicitly capture the interactions among actors and predict multiple future outcomes with calibrated uncertainty.
  \item Our planning module considers multiple possibilities of how the future
might unroll, and outputs a safe trajectory for the self-driving car that respects the laws of traffic and is compliant with other actors. 
  \item We address the costly probabilistic inference with a sample-based framework.
\end{itemize}
DSDNet conducts efficient inference based on message passing over a sampled set of continuous trajectories to obtain the future motion predictions.
It then employs a structured motion planning cost function, which combines a cost learned in a data-driven manner and a cost inspired by human prior knowledge on driving (\eg, traffic rules, collision avoidance) to ensure that the SDVs planned path is safe.
We refer the reader to Fig.~\ref{fig:model} for an overview of our full model.

We demonstrate the effectiveness of our model on two large-scale real-world datasets: \textbf{nuScenes} \cite{nuscenes2019} and \textbf{ATG4D}, as
well as one simulated dataset \textbf{CARLA-Precog} \cite{dosovitskiy2017carla,rhinehart2019precog}. Our method significantly outperforms previous
state-of-the-art results on both prediction and planning tasks.


\section{Related Work}


\paragraph{Motion Prediction:}
Two of the main challenges of prediction are modeling interactions among actors and making accurate multi-modal predictions.
To address these,
\cite{alahi2016social,gupta2018social,lee2017desire,rhinehart2019precog,tang2019multiple,zhao2019multi,casas2019spatially,interacttransform2020,ma2017forecasting}
learn per-actor latent representations and
model interactions by communicating those latent representations among actors. 
These methods can naturally work with VAE \cite{kingma2013auto} and produce multi-modal predictions.
However, they typically lack interpretability and it is hard to encode
prior knowledge, such as the traffic participants' desire to avoid collisions.
Different from building implicit distributions with VAE, \cite{chai2019multipath,lgn} build explicit distributions using mixture of modes (e.g., GMM) where it  is easier to perform efficient probabilistic inference. In this work,
we further enhance the model capacity with a non-parametric explicit distribution constructed over a dense set of trajectory samples. In concurrent
work \cite{phan2020covernet} use similar representation to ours, but they do not model social interactions and do not demonstrate how such a representation can
benefit planning.

Recently, a new prediction paradigm of performing joint detection and prediction has been proposed
\cite{zeng2019end,liang2020pnpnet,v2v,luo2018fast,pmlr-v87-casas18a,ilvm,priorpred}, in which actors' location
information is not known a-priori, and needs to be inferred from the sensors. In this work, we will demonstrate our approach in both settings:
using sensor data or history of actors' locations as input.

\paragraph{Motion Planning:}
Provided with perception and prediction results, planing is usually formulated as a cost minimization problem over trajectories. The cost function can be either manually engineered to guarantee certain properties
\cite{buehler2009darpa,fan2018baidu,montemerlo2008junior,ziegler2014trajectory}, or learned from data through imitation learning or inverse reinforcement
learning \cite{sadat2019jointly,wulfmeier2015maximum,zeng2019end,ziebart2008maximum}. However, most of these planners assume detection and prediction to be accurate and certain,
which is not true in practice. Thus, \cite{bandyopadhyay2013intention,hardy2013contingency,ppp,zhan2016non} consider
uncertainties in other actors' behaviors, and formulate collision avoidance in a probabilistic manner. Following this line of work, we also
conduct uncertainty-aware motion planning.

End-to-end self-driving methods  try to fully utilize the power of data-driven approaches and enjoy simple inference.
They typically use a neural network to directly map from raw sensor data to
 planning outputs, and are learned through imitation learning \cite{bojarski2016end,pomerleau1989alvinn}, or reinforcement
learning~\cite{codevilla2018end,muller2018driving} when a simulator~\cite{dosovitskiy2017carla,manivasagam2020lidarsim} is available. 
However, most of them lack interpretability and do not explicitly ensure safety. While our method also benefits from the power of deep learning, in contrast to the aforementioned approaches, we explicitly model interactions
between the SDV and the other dynamic agents, achieving safer planning. Furthermore, safety is explicitly accounted for in our planning cost functions. 

\paragraph{Structured models and Belief Propagation:}
To encode prior knowledge, there is a recent surge of deep structured
models{~\cite{belanger2016structured,chen2015learning,graber2018deep,schwing2015fully,ma2019deep}, which
use deep neural networks (DNNs) to provide the energy terms of a probabilistic graphical models (PGMs). 
Combining the powerful learning capacity of DNNs and the task-specific structure imposed by PGMs, deep structured models have been successfully applied to various computer vision problems, \eg, semantic segmentation~\cite{schwing2015fully}, anomaly detection~\cite{zhai2016deep}, contour segmentation~\cite{marcos2018learning}.
However, for continuous random variables,  inference is  very challenging. 
Sample-based belief propagation
(BP)~\cite{weiss2010belief,yedidia2003understanding,ihler2009particle,sudderth2010nonparametric,yamaguchi2014efficient,yamaguchi2012continuous}, address this issue by first constructing the approximation of the continuous distribution via Markov Chain Monte Carlo (MCMC) samples and then performing inference via BP. 
Inspired by these works, we design a deep structured model that can learn complex human behaviors from large data while incorporating our prior
knowledge. We also bypass the difficulty in continuous variable inference using
a physically valid sampling procedure.

\newcommand{\bs}{\mathbf{s}}
\newcommand{\bw}{\mathbf{w}}
\newcommand{\bpi}{\pmb{\pi}}
\newcommand{\btau}{\pmb{\tau}}
\newcommand{\cS}{\mathcal{S}}
\newcommand{\bbR}{\mathbb{R}}
\newcommand{\bF}{\mathbf{F}}
\newcommand{\bX}{\mathbf{X}}
\newcommand{\ba}{\mathbf{a}}
\newcommand{\bb}{\mathbf{b}}

\begin{figure}[t]
\begin{center}
  \includegraphics[height=3.3cm]{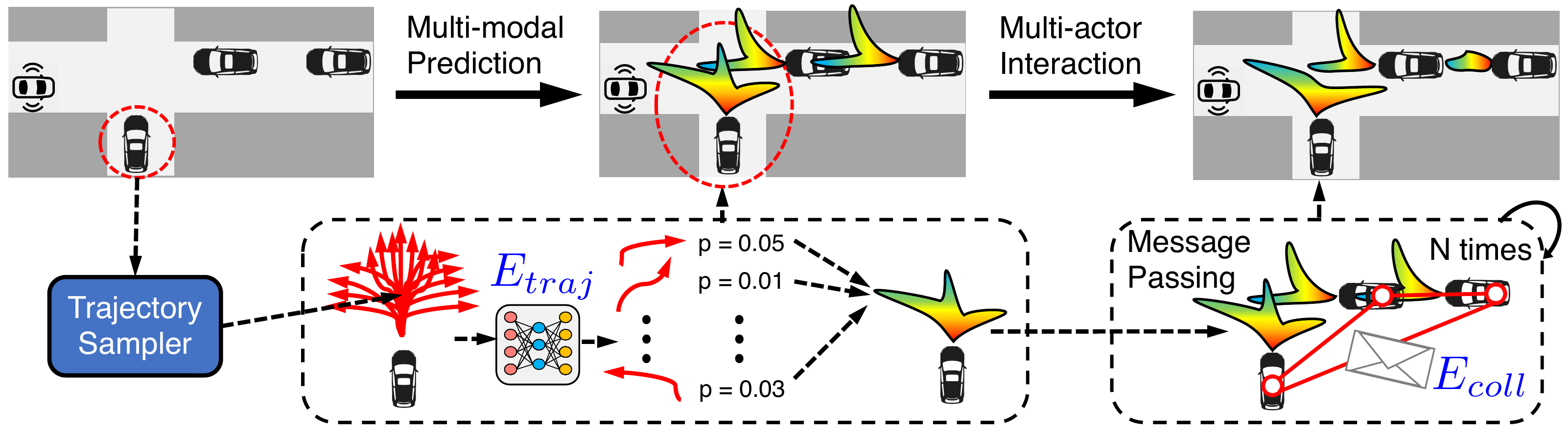}
\end{center}
\caption{\textbf{Details of the multimodal social prediction module:} For each actor, we sample a set of physically valid
trajectories, and use a neural network $E_{traj}$ to assign energies (probabilities) to them. To make different actors' behaviors socially consistent, we employ
message passing steps which explicitly model interactions and can encode human prior knowledge (collision avoidance). The final predicted
socially-consistent distribution is shown on top right.}
\label{fig:pred}
\end{figure}

\section{Deep Structured self-Driving Network}

Given  sensor measurements  and a map of the environment, the objective of a self-driving vehicle (SDV) is to select a trajectory to execute (amongst
all feasible ones) that is safe, comfortable, and allows the SDV to reach its destination.  
In order to plan a safe maneuver, a self-driving vehicle has to first understand its surroundings as well as predict how the future might evolve. 
It should then plan its motion by considering all possibilities of the future  weighting them properly. This is not trivial as the future is very
multi-modal and actors  interact with each other. Moreover, the inference procedure needs to be performed in a fraction of a second in order to have practical value.

In this paper we propose DSDNet, a single neural network that jointly detects actors in the scene, predicts a socially consistent multimodal distribution over their future behaviors, and produces  safe motion plans for the SDV. 
Fig.~\ref{fig:model} gives an overview of our proposed approach. We first utilize a backbone network to
compute the intermediate feature-maps, which are then used for detection, prediction and planning. After detecting actors with a
detection header, a deep structured probabilistic inference module computes the distributions of actors' future trajectories, taking into account the
interactions between them. 
Finally, our planning module outputs the planned trajectory by considering both the contextual information encoded in the feature-maps as well
as possible futures predicted from the model.

In the following, we first briefly explain the input representation, backbone network  and detection module. 
We then introduce our novel probabilistic prediction and motion planning  framework in
sections \ref{sec:method_prediction} and  \ref{sec:method_planning} respectively.
Finally, we illustrate how to train our model end-to-end in section \ref{sec:method_learning}.

\subsection{Backbone Feature Network and Object Detection}

Let $\bX$ be the  LiDAR point clouds and the HD map given as input to our system. 
Since LiDAR point clouds can be very sparse and the actors' motion  is an important cue for detection and prediction, we use the past 10 LiDAR sweeps (e.g., 1s of measurements) and 
voxelize them into a 3D tensor  \cite{luo2018fast,yang2018pixor,zhou2018voxelnet,zeng2019end}. We 
utilize HD maps as they provide a strong  prior  about the scene.
Following \cite{zeng2019end}, we rasterize the lanes with different semantics (\eg, straight, turning, blocked by traffic light) into different channels 
and concatenate them with the 3D LiDAR tensor to form our input representation. 
We then process  this 3D tensor with a deep convolutional network backbone and compute a backbone feature map {$\bF \in \bbR^{H \times W
\times C}$, where $H, W$ correspond to the spatial resolution after downsampling (backbone) and $C$ is the channel number.
We then employ a single-shot detection header on this feature map to output detection bounding boxes for the  actors in the scene.
We  apply two Conv2D layers separately on $\bF$, one for classifying if a location is occupied by an actor, the other for regressing the position offset, size, orientation and speed of each actor.
Our prediction and planning modules will then take these detections and the feature map as input to produce both a distribution over the actors'
behaviors and a safe planning maneuver. 
For more details on our detector and backbone network please refer to the supplementary material.

\subsection{Probabilistic Multimodal Social Prediction}
\label{sec:method_prediction}
In order to plan a safe  maneuver, we need to predict how other actors could potentially behave in the next few seconds. 
As actors move on the ground, we represent their possible future behavior using a trajectory defined as a sequence of 2D waypoints on birds eye view (BEV) sampled at $T$ discrete timestamps. 
Note that $T$ is the same duration as our planning horizon, and we compute the motion prediction distribution and a motion plan each time a new sensor measurement arises (i.e., every 100ms). 

\paragraph{Output Parameterization:}
Let $\bs_i \in  \bbR^{T \times 2}$ be the future trajectory of the $i$-th actor. 
We are interested in modeling the  joint distribution of all actors condition on the input, that is $p(\bs_1, \cdots, \bs_N | \bX)$. 
Modeling this joint distribution and performing efficient inference is challenging, as each actor has a high-dimensional continuous action space. 
Here, we propose to approximate this high-dimensional continuous space with a finite number of samples, and construct a non-parametric
distribution over the sampled space. Specifically, for each actor, we randomly sample $K$ possible future trajectory $\{\hat{\bs}_i^1, \cdots, \hat{\bs}_i^K\}$ from the original continuous trajectory space
$\bbR^{T \times 2}$.
We then constrain the possible future state of each actor to be one of those $K$ samples. 
To ensure samples are always diverse, dense\footnote{We would like the samples to cover the original continuous space and 
have high recall wrt the ground-truth future trajectories.} and physically plausible,
we follow the \textbf{N}eural
\textbf{M}otion \textbf{P}lanner (NMP) ~\cite{zeng2019end} and use a combination of straight, circle, and clothoid curves. 
More details and analysis of the sampler can be found in the supplementary material. 

\paragraph{Modeling Future Behavior of All Actors:} We employ an energy formulation to measure the probability of  each  possible future configuration of
all actors in the scene:
a configuration $(\bs_1, \cdots, \bs_N)$ has low energy if it is likely to happen.
We can then compute the  joint distribution of all actors' future behaviors as
\begin{align}
\label{eq:mrf_prob}
p(\bs_1,\cdots,\bs_N |\bX, \bw) = \frac{1}{Z} \exp\left( - E(\bs_1,\cdots,\bs_N |\mathbf{X}, \bw) \right),
\end{align}
where 
$\bw$ are learnable parameters, $\bX$ is the raw sensor data and $Z$ is the partition function {$Z=\sum
\exp(-E(\hat{\bs}_1^{k_1},\cdots,\hat{\bs}_N^{k_N}))$ summing over all actors' possible states}.

We construct the energy $E(\bs_1, \cdots, \bs_N | \bX, \bw)$ inspired by how humans drive, e.g., following common sense as well as traffic rules. 
For example, humans drive smoothly along the road and avoid collisions with each other.
Therefore, we decompose the energy $E$ into two terms.  The first term encodes the goodness of a future trajectory (independent of other
actors) while the second term explicitly encodes the fact that pairs of actors should not collide in the future. 
\begin{align}
\label{eq:mrf_energy}
E(\bs_1,\cdots,\bs_N | \mathbf{X}, \bw) = \sum_{i=1}^N E_{traj}(\bs_i | \mathbf{X}, \bw_{traj}) + \sum_{i=1}^N \sum_{i\neq j}^N E_{coll}(\bs_i, \bs_j
| \mathbf{X}, \bw_{coll})
\end{align}
where $N$ is the number of detected actors  and $\bw_{traj}$ and $\bw_{coll}$ are the parameters. 

Since the goodness $E_{traj}(\bs_i | \mathbf{X}, \bw_{traj})$ is hard to define manually, we use a neural network to learn it from data (see Fig.
\ref{fig:header}).
Given the sensor data $\bX$ and a proposed trajectory $\bs_i$, the network will output a scalar value.
Towards this goal, we first use the detected bounding box of the $i$-th actor and apply ROIAlign to the backbone feature map, followed by several convolution layers to compute the actor's feature.
Note that the backbone feature map is expected to encode rich information about both the environment and the actor.
\begin{wrapfigure}{r}{0.34\textwidth}
  \centering
    \includegraphics[width=0.32\textwidth]{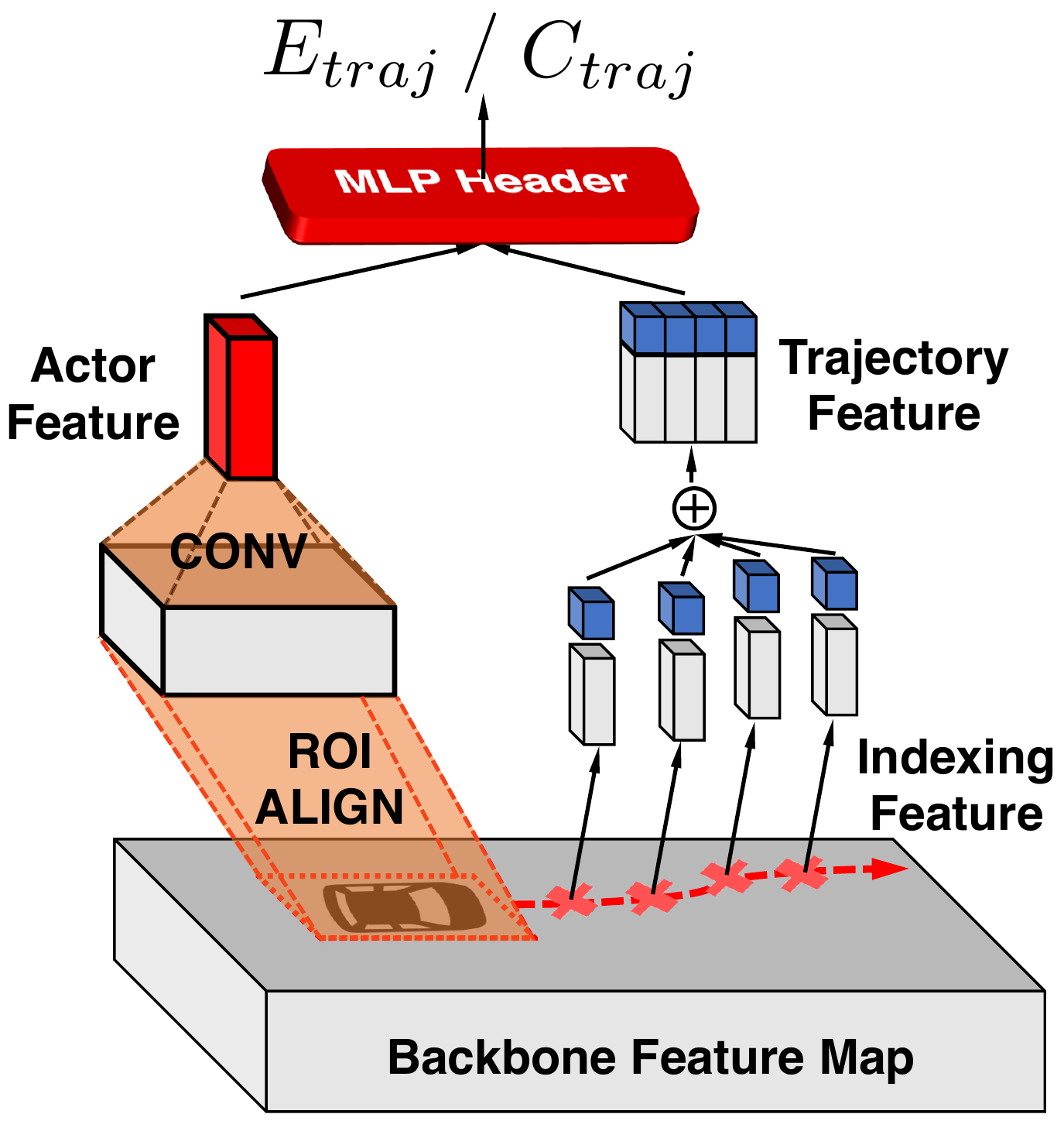}
  \caption{Neural header for evaluating $E_{traj}$ and $C_{traj}$.}
  \label{fig:header}
\end{wrapfigure}
We then index (bilinear interpolation) 
$T$ features on the backbone feature map at the positions of trajectory's waypoints,
and concatenate them together with $(x_t, y_t, cos(\theta_t), sin(\theta_t), distance_t)$ to form the trajectory feature of $\bs_i$. 
Finally, we feed both actor and trajectory features into an MLP which outputs $E_{traj}(\bs_i | \mathbf{X}, \bw_{traj})$. Fig.. \ref{fig:header}
shows an illustration of the architecture.

We use a simple yet effective collision
energy: $E(\bs_i, \bs_j) = \gamma$ if $\bs_i$ collides with $\bs_j$, and $E(\bs_i, \bs_j) = 0$ otherwise, to explicitly model the collision avoidance
interaction between actors as explained in the next paragraph.
We found this simple pairwise energy empirically works well, the exploration of other learnable pairwise energy is thus left as future work.

\paragraph{Message Passing Inference:}

For safety, our motion planner needs to take all possible actor's future into consideration. Therefore, motion forecasting needs to infer the probability of each actor taking a particular future trajectory: $p(\bs_i = \hat{\bs}_i^k | \bX, \bw)$. We thus  conduct marginal inference over the joint distribution. 
Note that the joint probability defined in Eq. (\ref{eq:mrf_energy}) represents a deep structured model (i.e., a Markov random field with potentials computed with deep neural networks). 
We utilize sum-product message passing  \cite{yedidia2003understanding} to estimate the marginal distribution per actor, taking into account the effects of all other actors by marginalization.
The marginal $p(\bs_i | \bX, \bw)$ reflects the uncertainty and  multi-modality in an actor's future behavior and will be leveraged by our planner.
We use the following update rule in an iterative manner for each actor ($\bs_i$):
\begin{align}
\label{eq:bp}
m_{ij}(\bs_j) & \propto \sum_{\bs_i \in \{\bs_i^k\}} e^{- E_{traj}(\bs_i) - E_{coll}(\bs_i, \bs_j) } \prod_{n \neq i,j} m_{ni}(\bs_i)
\end{align}
where $m_{ij}$ is the message sent from actor $i$ to actor $j$ and $\propto$ means equal up to a normalization constant. 
Through this message passing procedure, actors communicate with each others their  future intentions $\bs_i$ and how probable those intentions
are $E_{traj}(\bs_i)$. The collision energy $E_{coll}$ helps to coordinate intentions from different actors such that the behaviors are compliant and do not result in collision.
After messages have been passed for a fixed number of iterations, we compute the approximated marginal as 
\begin{align}
\label{eq:marginal}
p(\bs_i = \hat{\bs}_i^k | \bX, \bw) \propto e^{- E{traj}(\hat{\bs}_i^k) } \prod_{j \neq i} m_{ji}(\hat{\bs}_i^k).
\end{align}
Since  we typically have a small graph (less than 100 actors in a scene) and each $\bs_i$ only has $K$ possible values $\{\hat{\bs}_i^1, \cdots, \hat{\bs}_i^K\}$,  we can efficiently
evaluate Eq. (\ref{eq:bp}) and Eq. (\ref{eq:marginal}) via matrix multiplication on GPUs. In practice we find that our energy in Eq.
(\ref{eq:mrf_energy}) usually results in sparse graphs: most actor will only interact with nearby actors, especially the actors in the front and in
the back. As a result, the message passing converges  within 5 iterations 
\footnote{Although the sum-product algorithm is only exact for tree structures, it is shown to work  well in practice for graphs with
cycles~\cite{murphy1999loopy,weiss2010belief}.}.
With our non-highly-optimized implementation, the prediction module takes
less than 100 ms on average, and thus it satisfies our real-time requirements.

\subsection{Safe Motion Planning under Uncertain Future}
\label{sec:method_planning}
The motion planning module fulfills our final goal, that is, navigating  towards a destination while avoiding collision and obeying  traffic rules. 
Towards this goal, we build a cost function $C$, which assigns lower cost values to ``good" trajectory proposals and higher values to ``bad" ones.
Planning is then performed by finding the optimal trajectory with the minimum cost 
\begin{align}
  \label{eq:plan_def}
  \btau^* = \arg \min_{\btau \in \mathcal{P}} C\left(\btau | p(\bs_1,\cdots,\bs_N), \bX, \bw\right),
\end{align}
with $\btau^*$ the planned optimal trajectory   and $\mathcal{P}$  the set of physically realizable trajectories that do not violate the SDV's
dynamics.
In practice, we sample a large number of future trajectories for the SDV conditioned on its current dynamic state (\eg, velocity and acceleration) to
form $\mathcal{P}$, which gives us a finite set of feasible trajectories $\mathcal{P} = \{\hat{\btau}^1, \cdots,
\hat{\btau}^K\}$. 
We use the same sampler as described in section \ref{sec:method_prediction} to ensure we get a wide variety of physically possible  trajectories. 

\paragraph{Planning Cost:}
Given a  SDV trajectory $\btau$, we compute the cost  
based on how good $\btau$ is 1) conditioned on the scene, (\eg, traffic lights and road topology); 2) considering all other actors' future behaviors  (\ie, marginal
distribution estimated from the prediction module). 
We thus define our cost as
{\small  
\begin{align}
\label{eq:mp_cost}
C(\btau | p(\bs_1,\cdots,\bs_N), \bX, \bw) =  C_{traj}(\btau | \bX, \bw) + \sum_{i=1}^N\mathbb{E}_{p(\bs_i| \mathbf{X}, \bw)}\left[  C_{coll}(\btau, \bs_i |
\mathbf{X}, \bw)\right],
\end{align}}
where $C_{traj}$ models the goodness of a SDV trajectory   using a neural network. Similar to $E_{traj}$ in the prediction module, we use the trajectory
feature and ROIAlign extracted from the backbone feature map to compute this scalar cost value. 
The collision cost $C_{coll}$ is designed for guaranteeing safety and avoid collision: \ie, $C_{coll}(\btau, \bs_i) = \lambda$ if $\btau$ and $\bs_i$ colide, and $0$ otherwise. 
This ensures a dangerous trajectory $\btau$ will incur a very high cost and will be rejected by our cost minimization inference process.
Furthermore, Eq. (\ref{eq:mp_cost}) evaluates the expected collision cost $\mathbb{E}_{p(\bs_i| \mathbf{X}, \bw)}[C_{coll}]$.
Such a formulation is helpful for safe motion planning since the future is uncertain and we need to consider all possibilities, properly weighted by how likely they are to happen.

\paragraph{Inference:}
We conduct exact minimization over $\mathcal{P}$. $C_{traj}$ is a neural network based cost and we can evaluate all $K$
possible trajectories with a single batch forward pass.
$C_{coll}$ can be computed with a GPU based collision checker. As a consequence,
we can efficiently evaluate $C(\btau)$ for all $K$ samples and select the trajectory with minimum-cost as our final planning result. 

\subsection{Learning}
\label{sec:method_learning}
We train the full model (backbone, detection, prediction and planning) jointly with a multi-class loss defined as follows
\begin{align}
  \mathcal{L} = \mathcal{L}_\text{planning} + \alpha \mathcal{L}_\text{prediction} + \beta \mathcal{L}_\text{detection}.
\end{align}
where $\alpha, \beta$ are constant hyper-parameters. Such a multi-task loss can fully exploit the supervision for each task and help the
training\footnote{We find that  using only $\mathcal{L}_{planning}$ without the other two terms prevents the model from learning reasonable detection
and prediction.}.
\paragraph{Detection Loss:} 
We employ a standard detection loss $\mathcal{L}_{detection}$, which is a sum of classification and regression loss.
We use a cross-entropy classification loss and assign an anchor's label based on its IoU with any actor.
The regression loss is a smooth $\ell_1$ between our model regression outputs and the ground-truth targets. Those targets include position, size, orientation and velocity. 
We refer the reader to
the supplementary material for more details.

\paragraph{Prediction Loss:} As our prediction module outputs a discrete distribution for each actor's behavior, we employ cross-entropy  between our discrete distribution  and the true target. as our  prediction loss. 
 Once we sampled $K$ trajectories per actor, this loss can be
regarded as a standard classification loss over $K$ classes (one for each trajectory sample). The target class is set to be the closest trajectory sample to the
ground-truth future trajectory (in $\ell_2$ distance).  

\paragraph{Planning Loss:}
We expect our model to assign lower planning costs to better trajectories (\eg, collision free, towards the goal), and higher costs to
bad ones. However, we do not have direct supervision over the cost. Instead,  we utilize a max-margin loss, using expert behavior as
positive examples and randomly sampled trajectories as negative ones. 
We set large margins for dangerous behaviors such as
 trajectories with collisions. This allows  our model to penalize dangerous behaviors more severely.
More formally, our planning loss is defined as
\begin{align}
  \mathcal{L}_\text{planning} = \sum_{data}\max_{k}\left(\left[C\left(\btau^{gt} | \bX\right) - C\left(\hat{\btau}^{k} | \bX\right) + d^{k} +
  \gamma^{k}\right]_{+}\right), \nonumber
\end{align}
where $[\cdot]_{+}$ is a ReLU function, $\btau^{gt}$ is the expert trajectory and $\hat{\btau}^{k}$ is the $k$-th  trajectory sample. We also define
$d^k$ as the $\ell_2$ distance between $\hat{\btau}^k$ and $\btau^{gt}$, and $\gamma^{k}$ is a constant positive penalty if $\hat{\btau}^{k}$ behave dangerously, \eg, $\gamma_\text{collision}$ if $\btau^{k}$ collides with another actor and 0 otherwise. 


\section{Experimental Evaluation}\label{sec:experiments}
We evaluate our model on all three tasks: detection, prediction, and  planning. 
We show results on two large scale  real-world self-driving datasets:
\textbf{nuScenes} \cite{nuscenes2019} and our in-house dataset \textbf{ATG4D}, as well as the \textbf{CARLA} simulated dataset
 \cite{dosovitskiy2017carla,rhinehart2019precog}. 
We show that \textbf{1)} our prediction module largely outperforms the state-of-the-art on public benchmarks and we demonstrate the benefits of explicitly modeling the interactions between actors.
\textbf{2)} Our planning module achieves the safest planning results and largely decreases the collision and lane violation rate, compared to  competing methods.
\textbf{3)} Although sharing a single backbone to speedup inference, our model does not sacrifice detection performance compared to the state-of-the-art. 
We provide datasets' details and implementation details in the supplementary material. 

\begin{table}[t]
  \centering
  \begin{subtable}{1.0\linewidth}\centering
  \begin{tabular}{ccccccc|ccccccc}
    \specialrule{.2em}{.1em}{.1em}
    \textbf{nuScenes} & \multicolumn{3}{c}{L2 (m)} & \multicolumn{3}{c|}{Col (\textperthousand)} & \textbf{ATG4D} & \multicolumn{3}{c}{L2 (m)} &
    \multicolumn{3}{c}{Col (\textperthousand) }\\
    Method & 1s & 2s & 3s & 1s & 2s & 3s & Method & 1s & 2s & 3s & 1s & 2s & 3s\\
    \hline
    Social-LSTM \cite{alahi2016social} & 0.71 & - & 1.85 & 0.8 & - & 9.6 & FaF \cite{luo2018fast} & 0.60 & 1.11 & 1.82 & - & - & - \\
    CSP \cite{deo2018convolutional} & 0.70 & - & 1.74 & 0.4 & - & 5.8 & IntentNet \cite{pmlr-v87-casas18a} & 0.51 & 0.93 & 1.52 & - & - & - \\
    CAR-Net \cite{sadeghian2018car} & 0.61 & - & 1.58 & 0.4 & - & 4.9 & NMP \cite{zeng2019end}  & 0.45 & 0.80 & 1.31 & 0.2 & 1.1 & 5.9 \\
    NMP \cite{zeng2019end} & 0.43 & 0.83 & 1.40 & \textbf{0.0} & 1.4 & 6.5 & & & & & & & \\
    \hline
    \rowcolor{grey} \textbf{DSDNet} & \textbf{0.40} & \textbf{0.76} & \textbf{1.27} & \textbf{0.0} & \textbf{0.0} & \textbf{0.2} & \textbf{DSDNet}& \textbf{0.43} &
    \textbf{0.75} & \textbf{1.22} & \textbf{0.1} & \textbf{0.1} & \textbf{0.2} \\

    \specialrule{.1em}{.05em}{.05em}
  \end{tabular}
  \caption{Prediction from raw sensor data: $\ell_2$ and Col (collision rate), lower the better. 
  }
  \label{table:pred_table_pnp}
  \end{subtable}

  \begin{subtable}{1.0\linewidth}\centering
  \small
  \begin{tabular}{ccccccc>{\columncolor{grey}}c}
    \specialrule{.2em}{.1em}{.1em}
    \textbf{CARLA}      & DESIRE\cite{lee2017desire} & SocialGAN\cite{gupta2018social} & R2P2\cite{rhinehart2018r2p2}  &
    MultiPath\cite{chai2019multipath} & ESP\cite{rhinehart2019precog} & MFP\cite{tang2019multiple} & \textbf{DSDNet}\\
  \hline
    Town 1 & 2.422 & 1.141 & 0.770 & 0.68 & 0.447 & 0.279 & \textbf{0.195} \\
    Town 2 & 1.697 & 0.979 & 0.632 & 0.69 & 0.435 & 0.290 & \textbf{0.213} \\
    \specialrule{.1em}{.05em}{.05em}
  \end{tabular}
  \caption{Prediction from ground-truth perception: minMSD with K = 12, lower the better.
  }
  \label{table:pred_table_gt}
  \end{subtable}
  \caption{Prediction performance on nuScenes, ATG4D and  CARLA}
  
  \label{table:pred_table}
\end{table}

\subsection{Multi-modal Interactive Prediction}

\paragraph{Baselines:} On CARLA, we compare  with the  state-of-the-art  reproduced and reported from
\cite{chai2019multipath,rhinehart2019precog,tang2019multiple}.
On nuScenes\footnote{Numbers are reported on official validation split, since there is no joint detection and prediction benchmark.}, we
compare our method against several powerful multi-agent prediction approaches reproduced and reported from \cite{casas2019spatially}\footnote{\cite{casas2019spatially} replaced the original encoder (taking the ground-truth detection and tracking as input) with a learned CNN that takes LiDAR as input  for a fair comparison.}: \textbf{Social-LSTM} \cite{alahi2016social}, \textbf{Convolutional Social Pooling (CSP)} \cite{deo2018convolutional} and 
\textbf{CAR-Net} \cite{sadeghian2018car}.
On ATG4D, we compare with LiDAR-based joint detection and prediction models, including \textbf{FaF}~\cite{luo2018fast} and 
\textbf{IntentNet}~\cite{pmlr-v87-casas18a}. We also  compare with \textbf{NMP}~\cite{zeng2019end} on both datasets.

\paragraph{Metrics:} Following previous works \cite{alahi2016social,gupta2018social,lee2017desire,luo2018fast,zeng2019end}, we report \textbf{L2
Error} between our prediction (most likely) and the ground-truth at different future
timestamps. 
We also report \textbf{Collision Rate}, defined as the percentage of actors that will collide with others if they follow the
predictions.
We argue that a socially consistent prediction model should achieve low collision rate, as avoiding collision is always one of the highest priorities
for a human driver.
On \textbf{CARLA}, we follow \cite{rhinehart2019precog} and use \textbf{minMSD} as our metric, which is the minimal mean squared distance between the  top 12 predictions
and the ground-truth. 

\paragraph{Quantitative Results:}
As shown in  Table~\ref{table:pred_table_pnp}, our method achieves the best results on both datasets. 
This is impressive as  most  baselines use $\ell_2$ as  training objective, and thus are directly favored by the $\ell_2$ error metric, while our approaches uses
cross-entropy loss  to learn proper distributions and capture multi-modality.
Note that multimodal techniques are thought to score worst in this metric (see e.g., \cite{chai2019multipath}). 
Here, we show that it is possible to model multi-modality while achieving lower $\ell_2$ error, as the model can better understand actors' behavior.
Our approach also significantly reduces the collisions between the actors' predicted trajectories, which justifies the benefit of our multi-agent interaction modeling. 
We further evaluate our prediction performance when assuming ground-truth perception / history are known, instead of predicting using noisy detections
from the model. We conduct this evaluation on CARLA where all previous methods use this settings. As shown in Table~\ref{table:pred_table_gt},
our method again significantly outperforms previous best results.

\begin{table*}[t]
\centering
\begin{tabular}{cccccccccc}
  \specialrule{.2em}{.1em}{.1em}
  Model & \multicolumn{3}{c}{Collision Rate (\%)} & \multicolumn{3}{c}{Lane Violation Rate (\%)} & \multicolumn{3}{c}{L2 (m)}  \\
  & 1s &\  \ \ 2s \ \ \ & 3s &\ \ \ \  1s \ \  \ \  &\ \ \ 2s\ \ \ &\ \  \ \ 3s\ \  \ \ & 1s & 2s & 3s\\
  \hline
  Ego-motion & 0.01 & 0.54 & 1.81 & 0.51 & 2.72 & 6.73 & 0.28 & 0.90 & 2.02 \\
  IL & 0.01 & 0.55 & 1.72 & 0.44 & 2.63 & 5.38 & \textbf{0.23} & \textbf{0.84} & \textbf{1.92}\\
  Manual Cost& 0.02 & 0.22 & 2.21 & 0.39 & 2.73 & 5.02 & 0.40 & 1.43 & 2.99 \\
  Learnable-PLT~\cite{sadat2019jointly}& 0.00 & 0.13 & 0.83 & - & - & - & - & - & -\\
  NMP~\cite{zeng2019end} & 0.01 & 0.09 & 0.78 & 0.35 & 0.77 & 2.99 & 0.31 & 1.09 & 2.35\\
  \hline
  \rowcolor{grey} \textbf{DSDNet} & \textbf{0.00} & \textbf{0.05} & \textbf{0.26} & \textbf{0.11} & \textbf{0.57} & \textbf{1.55} & 0.29 & 0.97 & 2.02\\
  \specialrule{.1em}{.05em}{.05em}

\end{tabular}
\caption{{\bf Motion planning performance} on ATG4D. All metrics are computed in a cumulative manner across time,  lower the better.}
\label{table:plan_table}
\end{table*}

\begin{table}[t]
\centering
\begin{tabular}{ccccc|ccccc}
  \specialrule{.2em}{.1em}{.1em}
  \textbf{nuScenes} & \multicolumn{4}{c|}{Det AP @ meter} & \textbf{ATG4D} &
  \multicolumn{4}{c}{Det AP @ IoU}\\
  Method & 0.5 & 1.0 & 2.0 & 4.0 & Method & 0.5 & 0.6 & 0.7 & 0.8\\
  \hline
  Mapillary\cite{min2019multi} & 10.2 & 36.2 & 64.9 & 80.1 & FaF~\cite{luo2018fast} & 89.8 & 82.5 & 68.1 & 35.8\\
  PointPillars~\cite{lang2019pointpillars} & 55.5 & 71.8 & 76.1 & 78.6 & IntentNet~\cite{pmlr-v87-casas18a} & 94.4 & 89.4 & 78.8 & 43.5\\
  NMP~\cite{zeng2019end} & 71.7 & 82.5 & 85.5 & 87.0 & Pixor~\cite{yang2018pixor} & 93.4 & 89.4 & 78.8 & 52.2\\
  Megvii~\cite{zhu2019class} & \textbf{72.9} & 82.5 & 85.9 & \textbf{87.7} & NMP~\cite{zeng2019end} & \textbf{94.2} & \textbf{90.8} & \textbf{81.1} & 53.7 \\
  \hline
  \rowcolor{grey} \textbf{DSDNet} & 72.1 & \textbf{83.2} & \textbf{86.2} & 87.4 & \textbf{DSDNet} & 92.5 & 90.2 & \textbf{81.1} & \textbf{55.4}\\
  \specialrule{.1em}{.05em}{.05em}
\end{tabular}
\caption{\textbf{Detection performance}: higher  is better.
Note that although our method uses single backbone for multiple challenging tasks, our detection module can achieve on-par
performance with the state-of-the-art.
}
\label{table:det_table}
\end{table}

\subsection{Motion Planning}
\paragraph{Baselines:}
We implement multiple baselines for comparison, including both neural-based and classical planners:
\textbf{Ego-motion} takes past 1 second positions of the ego-car and use a 4-layer MLP to predict the future locations, as the ego-motion usually providse strong cues of how the SDV will move in the future.
\textbf{Imitation Learning (IL)} uses the same backbone network as our model 
but directly regresses an output trajectory for the SDV. We train such a regression model with $\ell_2$ loss w.r.t. the ground-truth planning trajectory.
\textbf{Manual Cost} is a classical sampling based motion planner based on a manually designed cost function
encoding collision avoidance and route following. The planned trajectory is chosen by finding the trajectory sample with  minimal cost. 
We also include previously published learnable motion planning methods: \textbf{Learnable-PLT} \cite{sadat2019jointly} and \textbf{Neural Motion Planner (NMP)} \cite{zeng2019end}. 
These two method utilize a similar max-margin planning loss as ours. However, \textbf{Learnable-PLT} only consider the most probable
future prediction, while \textbf{NMP} assumes planning is independent of prediction given the features. 


\paragraph{Metrics:}
We exploit three metrics to evaluate motion planning performance. 
\textbf{Collision Rate} and \textbf{Lane Violation rate} are the ratios of frames at which our planned trajectory either collides with other actors' ground-truth future
behaviors, or touches / crosses a lane boundary, up to a specific future timestamp. Those are important safety metrics (lower is better).  
\textbf{L2 to expert path} is the average $\ell_2$ distance between the planning trajectory and the expert driving path. 
Note that the expert driving path is just one among many possibilities, thus rendering this metric not perfect. 

\paragraph{Quantitative Results:}
The planning results are shown in Table~\ref{table:plan_table}. We can observe that: \textbf{1)} our proposed method provides the safest plans, as we achieve much lower collision  and lane violation rates compared to all other methods. 
\textbf{2)} Ego-motion and IL achieves the best $\ell_2$ metric, as they employ the power of neural networks and directly optimize the $\ell_2$ loss. However, they have high collision rate, which indicates directly mimicking expert demonstrations  is still insufficient to learn a safety-aware self-driving stack. 
In contrast, by learning interpretable intermediate results (detection and prediction) and by incorporating prior knowledge (collision
cost), our model can achieve much better results. The later point is further validated by comparing to NMP, which, despite learning detection and
prediction, does not explicitly condition on them during planning. 
\textbf{3)} Manual-Cost and Learnable-PLT explicitly consider collision avoidance and traffic rules. However, unlike our approach, they only take the most likely motion forecast  into consideration. 
Consequently, these methods have a higher collision rate than our approach.


\begin{table}[t]
  \centering
  \begin{subtable}{1.0\linewidth}\centering
  \begin{tabular}{cc|cc}
    \specialrule{.2em}{.1em}{.1em}
    Multi-modal\ \  & \ \ Interaction & Pred Col (\textperthousand) & Pred L2 (m) \\
    \hline
    &  & 4.9 & 1.28 \\
    \checkmark & & 2.4 & \textbf{1.22}  \\
    \checkmark & \checkmark & \textbf{0.2} & \textbf{1.22} \\
    \specialrule{.1em}{.05em}{.05em}
  \end{tabular}
  \caption{Prediction ablation studies on multi-modality and pairwise interaction modeling.
  }
  \label{table:ablation_pred}
  \end{subtable}
  \begin{subtable}{1.0\linewidth}\centering
  \begin{tabular}{l|cc}
    \specialrule{.2em}{.1em}{.1em}
    w/ Prediction (Type) & Plan Col (\%) & Lane Vio (\%) \\
    \hline
    N/A & 0.60 & 1.64 \\
    most likely & 0.45 & 1.60 \\
    multi-modal & \textbf{0.26} & \textbf{1.55} \\
  \specialrule{.1em}{.05em}{.05em}
  \end{tabular}
  \caption{Motion planning ablation studies on incorporating different prediction results. 
  }
  \end{subtable}
  \caption{Ablation Study for prediction and planning modules}
  \label{table:ablation_plan}
  \label{table:ablation_table}
\end{table}

\subsection{Object Detection Results}
We show our object detection results on \textbf{nuScenes} and \textbf{ATG4D}.
Although we use a single backbone for all three challenging tasks, we show in Table \ref{table:det_table} that our model can achieve 
similar or better results than   state-of-the-art LiDAR detectors on both datasets.
On \textbf{nuScenes}\footnote{We conduct the comparison on the official validation split, as our model currently only focuses on vehicles while the
testing benchmark is built for multi-class detection.},
we follow the official benchmark and use detection average precision (AP) at different distance (in meters) thresholds as our metric.
Here Megvii~\cite{zhu2019class} is the leading method on the leaderboard at the time of our submission. We use a smaller resolution (Megvii has 1000
pixels on each side while ours is 500) for faster online inference, yet achieve on-par performance. We also conduct experiments on ATG4D. 
Since our model uses the same backbone as Pixor~\cite{yang2018pixor}, which only focuses on detection, we demonstrate that a multi-task formulation does not sacrifice detection performance.

\subsection{Ablation Study and Qualitative Results}

Table~\ref{table:ablation_pred} compares different prediction modules with
the same backbone network. We can see that explicitly modeling a multi-modal future  significantly boosts the prediction performance in terms of both
collision rate and $\ell_2$ error, comparing to a deterministic (unimodal) prediction. The performance is  further boosted if the prediction module explicitly models the future interaction between multiple actors,  particularly in collision rate.
Table~\ref{table:ablation_plan} compares motion planners that consider different prediction results. We can see that explicitly incorporating
future prediction, even only the most likely prediction, will boost the motion planning performance, especially the collision rate.  
Furthermore, if the motion planner takes multi-modal futures into consideration, it achieves the best performance among the three. This further
justifies our model design.

\begin{figure*}[t]
\centering
\setlength{\tabcolsep}{1pt}
\begin{tabular}{ccc}
&\includegraphics[width=0.32\linewidth]{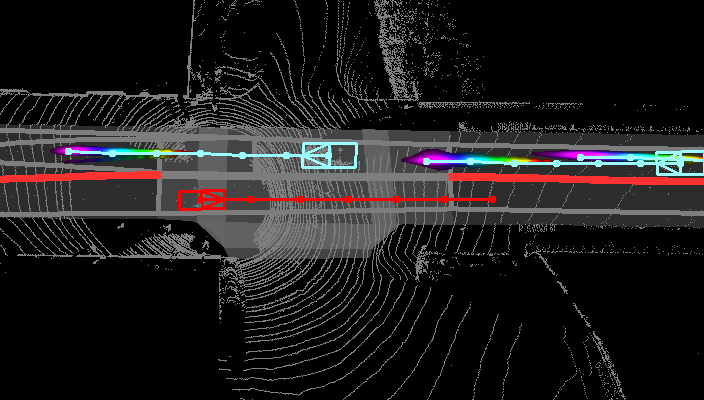}
\includegraphics[width=0.32\linewidth]{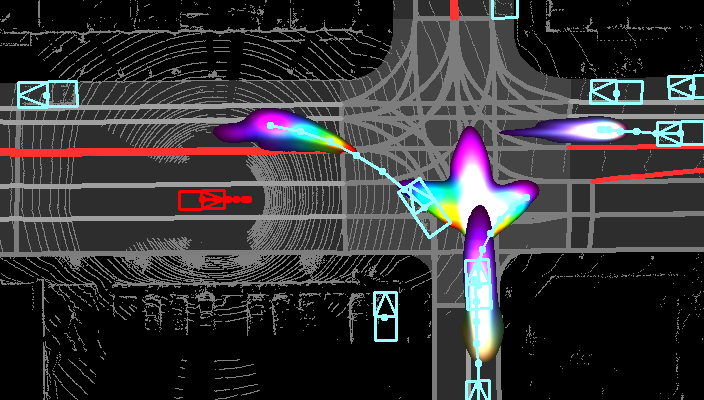}
\includegraphics[width=0.32\linewidth]{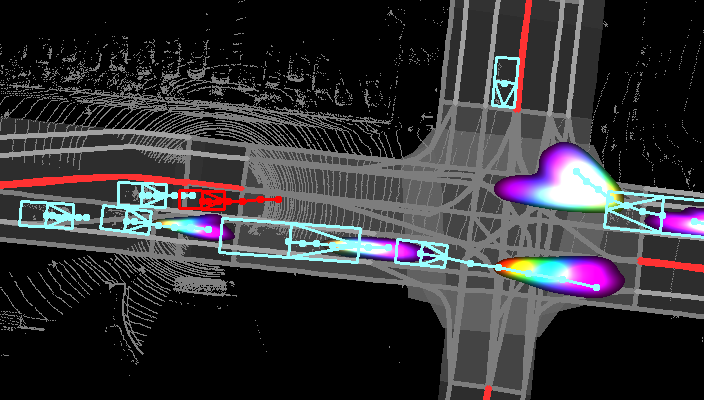}\\

&\includegraphics[width=0.32\linewidth]{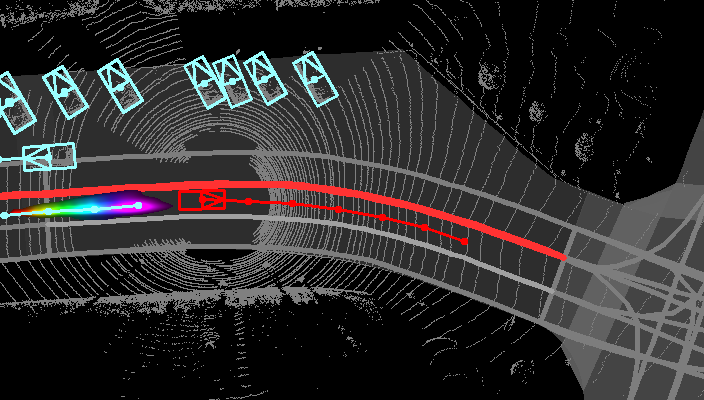}
\includegraphics[width=0.32\linewidth]{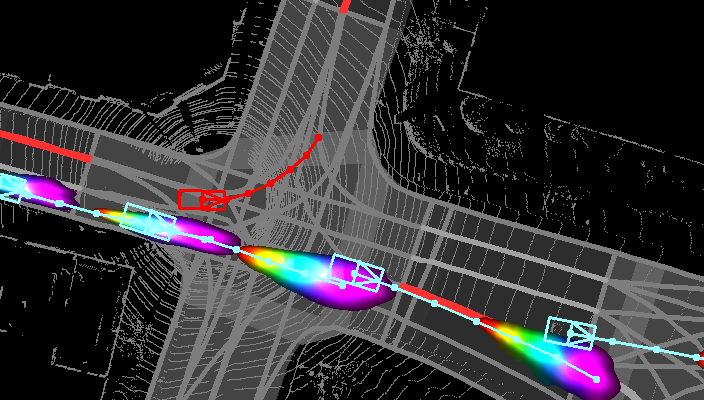}
\includegraphics[width=0.32\linewidth]{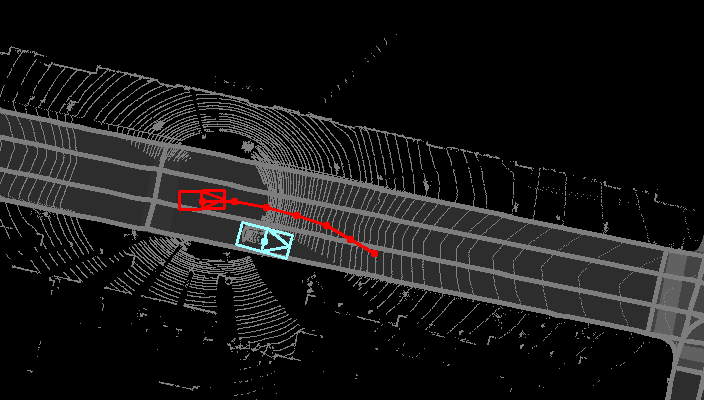}\\

\end{tabular}
\caption{Qualitative results on ATG4D. Our prediction module can capture multi-modalities when vehicles approach  intersections, while
  being certain and accurate when vehicles  drive along a single lane (top). Our model can produce smooth trajectories which follow the lane
  and are compliant with other actors (bottom). Cyan boxes: detection. Cyan trajectory: prediction. Red box: ego-car. Red trajectory: our planning.
We overlay the predicted marginal distribution for different timestamps with different colors  and only show high-probability regions.}
\label{fig:vis}
\end{figure*}

We show  qualitative results in Fig.~\ref{fig:vis}, where we visualize our detections, predictions, motion planning trajectories, and the predicted
uncertainties. We use different colors for different future timestamps to visualize high-probability actors' future positions estimated from our
prediction module. Thus larger `rainbow' areas mean more uncertain.
On the first row, we can see the predictions are certain when vehicles drive along the lanes (left), while we see multi-modal predictions when vehicles
approach  an intersection (middle, right).
On the second row, we can see our planning can nicely follow the lane (left), make a smooth left turn (middle), and take a nudge when an obstacle is blocking
our path (right).

\section{Conclusion}
In this paper, we propose DSDNet,
which is built on top of a single backbone network that takes as input raw sensor data and an HD map and performs perception, prediction, and planning under a unified framework.
In particular, we build a deep energy based model to parameterize the joint distribution of future trajectories of all traffic  participants.
We resort to a sample based formulation, which enables  efficient inference of the marginal distribution via belief propagation.
We design a structured planning cost which encourages traffic-rule following and collision avoidance for the SDV.
We show that our model outperforms the state-of-the-art on several challenging datasets.
In the future, we plan to explore more types of structured energies and cost functions. 

\clearpage
\section*{Acknowledgement}
We would like to thank Ming Liang, Kelvin Wong, Jerry Liu, Min Bai, Katie Luo and Shivam Duggal
for their helpful comments on the paper.

%
%
\bibliographystyle{splncs04}
\bibliography{mybib}
\clearpage

\section*{Supplementary Materials}
\appendix


\section{Datasets}
\label{supp:data}
\paragraph{CARLA:} This  is a public available multi-agent trajectory prediction dataset, collected by
\cite{rhinehart2019precog} using CARLA simulator \cite{dosovitskiy2017carla}. It contains over 60k training sequences, 7k testing sequences
collected from Town01, and 17k testing sequences from Town02. Each sequence is composed of 2 seconds of history, and 4 seconds future information.

\paragraph{nuScenes:} It contains 1000 driving snippets of length 20 seconds each. 
LiDAR point clouds are collected at 20Hz, and labels 3D bounding boxes are provided at  2Hz. 
To augment the labels, we generate bounding boxes for non-labeled frames using linear interpolation from 2 consecutive labeled frames. 
Since nuScenes dataset currently does not provide routing information for motion planning, we only conduct detection and prediction studies.
We follow the official data split and compare against other methods on the ``car" class.

\paragraph{ATG4D:} 
We  collected a challenging self-driving datasets over multiple cities across North America. 
It contains $\sim$ 5,000 snippets collected from 1000 different trips, with a 64-beam LiDAR running at 10 Hz and HD maps. 
We also labeled the data at 10 Hz, with maximum labeling range of 100 meters. We ignore parking areas far from the roads, as they will not interact with the SDV. 
We split out 500 snippets for testing and evaluate the full autonomy stack including motion planning, prediction as well as detection performance.

\section{Network Architecture Details}
\label{supp:arch}
In the following, we first describe our backbone network, the detection header, as well as the header for computing  prediction ($E_{traj}$) and planning $C_{traj}$.
Note we use the same architecture for nuScenes and ATG4D, but a slightly different one for CARLA as the setting there is different, which we will
explain in section \ref{supp:detail}.

\paragraph{Backbone:}
Our backbone is adapted from the detection network of \cite{yang2018pixor,zeng2019end}, which has 5 blocks of layers in total. 
There are \{2, 2, 3, 6, 5\} Conv2D layers with   \{32, 64, 128, 256, 256\} number of filters in those 5 blocks respectively. 
All Conv2D kernels are 3x3  and have stride 1. 
For the first three blocks, we use  a max-pooling layer after each block to downsample the feature map by 2. 
After the 4-th block, we construct a multi-scale feature map by resizing the feature maps after each block to be of the same size (4 times smaller than the input) and then concatenate them together. 
This multi-scale feature map is then fed to the 5-th block. The final feature map computed by the 5-th block has a downsample rate of 4, and is  shared for detection, prediction,
and motion planning modules.

\paragraph{Detection Header:}
We  use feature maps from the backbone and apply a
single-shot detection header, similar to SSD~\cite{liu2016ssd}, to predict the location, shape, orientation and velocity of each actor. 
More specifically, the detection header contains two Conv2D layers with $1\times1$ kernel, one for classification and the other one for regression. 
We apply the two Conv2D on the backbone feature map separately. 
To reduce the variance of regression outputs, we follow SSD~\cite{liu2016ssd} and use a set of predefined anchor boxes: each pixel at the backbone
feature map is associated with 12 anchors, with different sizes and aspect ratios.
We predict a classification score $p_{i,j}^k$  for each pixel $(i, j)$ and anchor $k$ on the feature map, which indicates how likely it is for an
actor to be presented at this location. 
For the  regression layer, the header outputs the offset values at each location. 
These offset values include position offset $l_x$, $l_y$, size $s_w$, $s_h$, heading angle $a_{sin}$, $a_{cos}$, and velocity $v_x$, $v_y$. 
Their corresponding ground-truth target values can be computed using the labeled bounding box, namely,
$$ l_x = \frac{x^{l}-x^{a}}{w^{a}} \quad l_y = \frac{y^{l} - y^{a}}{h^{a}},$$
$$ s_w = \log \frac{w^{l}}{w^{a}} \quad s_h = \log \frac{h^{l}}{h^{a}},$$
$$ a_{sin} = \sin(\theta^{l} - \theta^{a}) \quad a_{cos} = \cos(\theta^{l} - \theta^{a}),$$
$$ v_x = l_x^{t=1} - l_x^{t=0} \quad v_y = l_y^{t=1}-l_y^{t=0},$$
where subscript $l$ means label value, and $a$ means anchor value.
Finally, we combine these two outputs and apply an NMS operation to determine the bounding boxes for all actors and their initial speeds. 

\paragraph{Prediction ($E_{traj}$) / Planning ($C_{traj}$) Headers:} In addition to the detection header, our model has two headers: one outputs prediction energy
$E_{traj}$, the other outputs motion planning costs $C_{traj}$. 
Note that these two headers have the same architecture, but different learnable parameters. 
After computing the backbone feature map, we apply four Conv2D layers with  $128$ filters. This increases the model
capacity to better handle multiple tasks.  
Then, to extract the actor features, we perform  ROI align based on the actor's detection bounding box, which
output a $16\times16\times128$ feature tensor for this actor. 
We then apply another four Conv2D layers, each with  a downsample rate of 2 and filter
size $\{256, 512, 512, 512\}$ respectively. This gives us a $512$ dimensional feature vector for each actor. 
Note that we parameterized trajectories with 7 waypoint (2Hz for 3 second, including the inital waypoint at 0 second). 
To extract trajectory features, we first
index the feature on our header feature map at those waypoints with bilinear interpolation. This gives us a $7 \times 128$ feature. We then concatenate this feature with the $(x_t,
y_t, {\cos}\theta_t, {\sin}\theta_t, \text{distance}_t)$ of those $7$ waypoints, where $(x_t, y_t)$ is the coordinate of that waypoint,
$({\cos}\theta_t, {\sin}\theta_t)$ is the direction, and $\text{distance}_t$ is the traveled distance along the trajectory up to that
waypoint. Finally, we feed the actor and trajectory features to a 5 layer MLP to compute the $E_{traj}$ / $C_{traj}$ value for this trajectory. The MLP has
$(1024, 1024, 512, 256, 1)$ neurons for each layer.

\section{Trajectory Sampler Details}
\label{supp:sampler}
Following NMP~\cite{zeng2019end}, we assume a bicycle dynamic model for vehicles, and we use a combination of straight line, circle arcs, and clothoid
curves to sample possible trajectories. More specifically, to sample a trajectory for a given detected actor, we first estimate its initial position
and speed as well as heading angle from our detection output. We then sample the mode of this trajectory, \ie, straight, circle, clothoid proportional to $(0.3,
0.2, 0.5)$ probability.  Next, we uniformly sample values of control parameters for the chosen mode, \eg, radius for circle mode, radius and canonical heading
angle of clothoid. These sampled parameters determine the shape of this trajectory. We then sample an acceleration value, and compute the velocity values for the next 3
seconds based on this acceleration and the initial speed. Finally, we go along our sampled trajectory  with our sampled velocity, to determine the
waypoints along this trajectory for the next 3 seconds. 

In our experiments, we notice that different numbers of samples used for inference will affect the final performance. For a metric only cares about
precision, e.g., L2, which we used on nuScenes and ATG4D, more samples generally produces better performance. For instance, increasing number of
trajectory samples (for inference) from 100 to 200 will decrease final timestep L2 error from 1.29m to 1.22m on ATG4D, but further increase number of
samples only brings marginal improvements. Due to the consideration of memory and speed, we use 200 trajectory samples during inference and 100
trajectory samples for training. However, for a metric that considers both diversity and precision, e.g., minMSD (CARLA), there is a sweet point of
number of samples. On CARLA, we found that sample 100 trajectories during inference performs worse than sampling 50 samples, which corresponds to 0.24
and 0.18 minMSD on validation set respectively, and sampling 1000 trajectories performs the worst, producing a minMSD of 0.30. This is because when
presented with a set of dense samples, trajectories are spatially very close to each other. As a result, the highest-scored samples and its nearby
samples will have very similar scores, and thus selected by the top-K evaluation procedure, which loses some diversity. We also found on CARLA, it's
helpful to regress a future trajectory from the backbone and add it to the trajectory samples set before our prediction module, in order to augment
the sampled set and alleviate any potential gaps between our dynamic model assumption and the dynamic model used in CARLA.

\section{Implementation Details}
\label{supp:detail}
\paragraph{nuScenes}
We follow the dataset range defined by the creators of nuScenes, and use an input region of $[-49.6, 49.6] \times [-49.6, 49.6] \times [-3, 5]$ meters centered at the ego car. We aggregate the
current  LiDAR sweep with past 9 sweeps (0.5s period), and voxelize the space with $0.2 \times 0.2 \times 0.25$ meter per voxel resolution. This gives us  a $496
\times 496 \times 320$ input LiDAR tensor. We further rasterize the map information with the same resolution. The map information includes road mask,
lane boundary and road boundary, which provide critical information for predicting the behavior of a vehicle.

We train our model for the car class, using Adam optimizer with initial learning rate of 0.001. We decay the learning rate  by 10 at the 6-th  and 7-th epoch respectively, and
stop  training at 8-th epoch. The training batch size is 80 and we use 16 GPUs in total. For detection, we treat anchors larger than 0.7 IoU with labels as positive examples, and smaller than 0.5 IoU
as negative, and treat others as ignore. We adopt hard-mining to get good detection performance. 
For training the prediction, we treat a detection as positive if it has larger than 0.1 IoU with labels, and
only apply prediction loss on those positive examples. 
Besides, we apply data augmentation~\cite{zhu2019class} during training: randomly translating a
frame (-1 to 1 meters in XY and -0.2 to 0.2 for Z), random rotating along the Z axis (-45 degree to 45 degree), randomly scaling the whole frame (0.95 to
1.05), and randomly flipping over the X and Y axes.

\paragraph{CARLA-PRECOG:} 
We train the model with Adam optimizer, using a learning rate of 0.0001, and decay by 10 at 20 epochs and 30 epochs, and finish
training at 40 epochs. We use batch size of 160 on 4 GPUs. No other data augmentation is applied on this dataset.

Since the dataset and experiment setup is different from nuScenes (Carla has no map and only 4 height channels for LiDAR; the prediction task is also
provided with ground-truth actors' locations), we use a slightly different architecture. We first rasterize the LiDAR and the past positions of all
actors similar to PRECOG~\cite{rhinehart2019precog}, and feed them into a shallower backbone network with 5 blocks of Conv2D layers, each has
\{2,2,2,2,2\} layers respectively. We then compute the feature of each actor by concatenating the actor features extracted from the backbone (ROIAlign
followed by a number of convolutional layers) and a motion feature.
Such a motion feature is computed by applying several Conv1D layers on top of the past locations of an actor, which is a $T \times 3$ tensor. Those
$3$ dimensions are x, y coordinates and timestamp indexes. Finally, we compute the $E_{traj}$ and prediction results using the same architecture we have
described in Section.~\ref{supp:arch}.

\section{More Qualitative Results}

Additionally we provide more qualitative results showing our inference results on ATG4D. 
In Figure.~\ref{fig:vis1} we show the inputs and outputs
of our model, and in Figure.~\ref{fig:vis2} we explain how we visualize the prediction uncertainty estimated by our model.
We can clearly see that our approach produces multimodal estimates  when an actor is approaching  an intersection. 
We provide more cases showing prediction results in Figure.~\ref{fig:vis3}. 
Finally, we show our motion planner can handle various situations including lane following, turning and nudging around other
vehicles to avoid collision in Figure.~\ref{fig:vis4}. Our detections and predictions are shown in cyan, and the ego-car as well as motion planning
results are shown in red.

\begin{figure*}[h]
\centering
\setlength{\tabcolsep}{1pt}
\begin{tabular}{ccc}
  \includegraphics[width=0.32\linewidth]{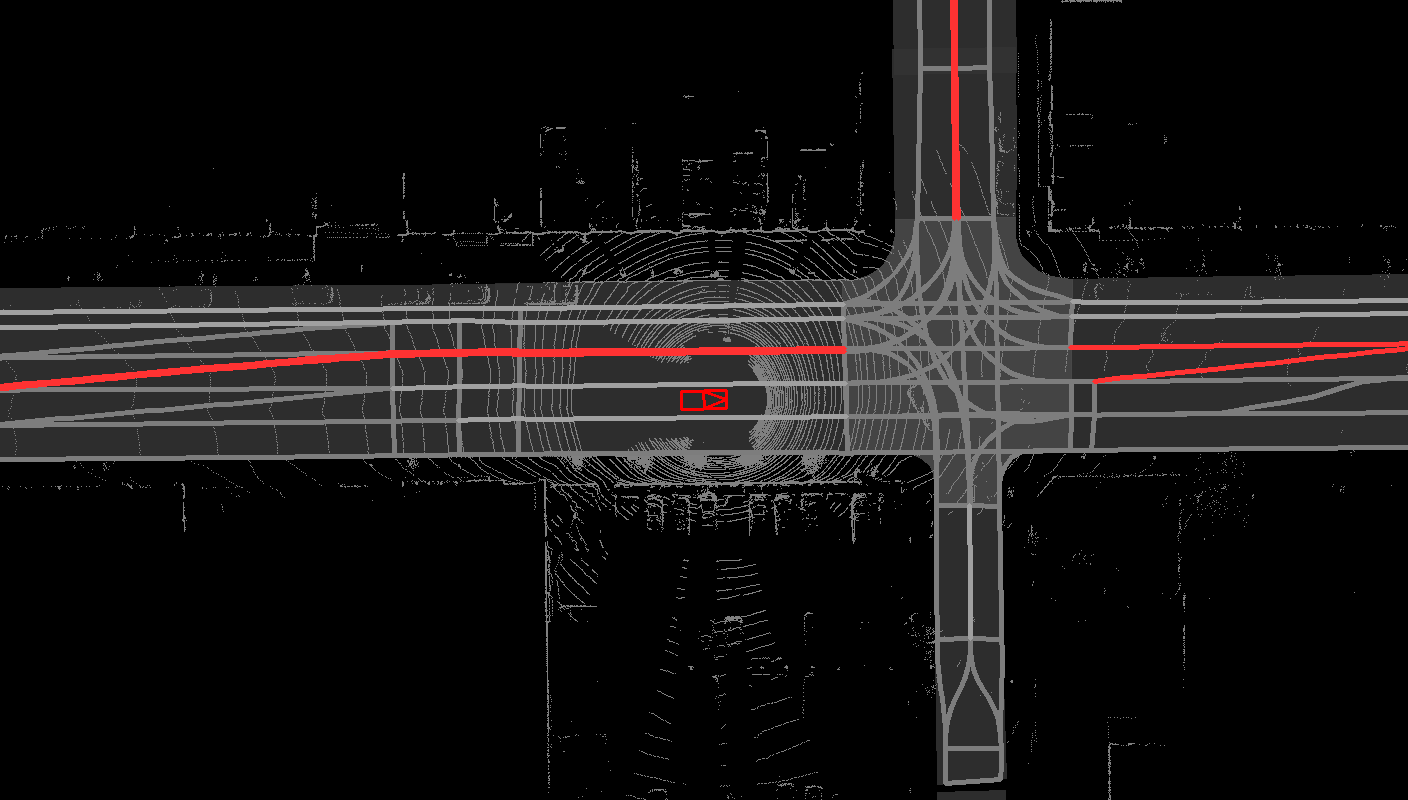}
  \includegraphics[width=0.32\linewidth]{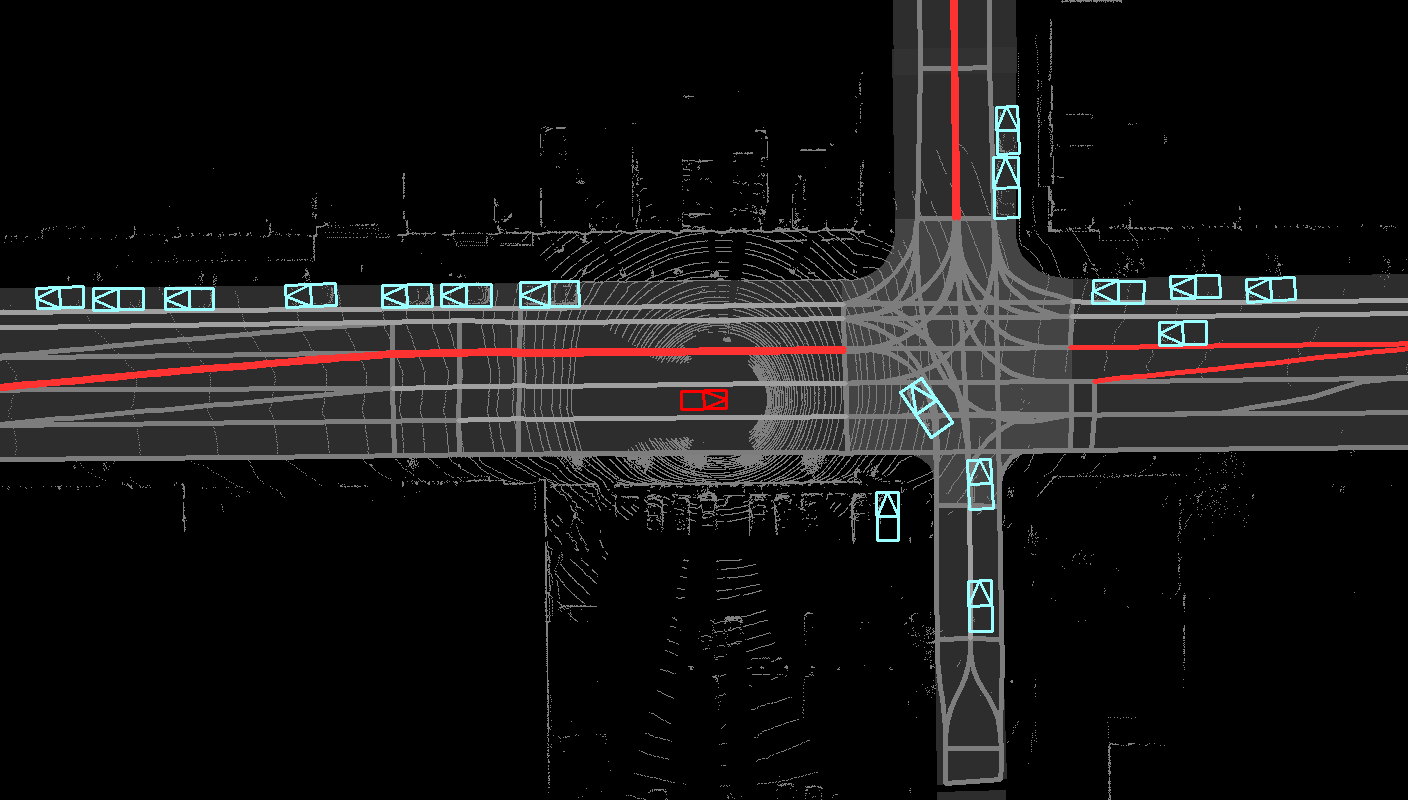}
  \includegraphics[width=0.32\linewidth]{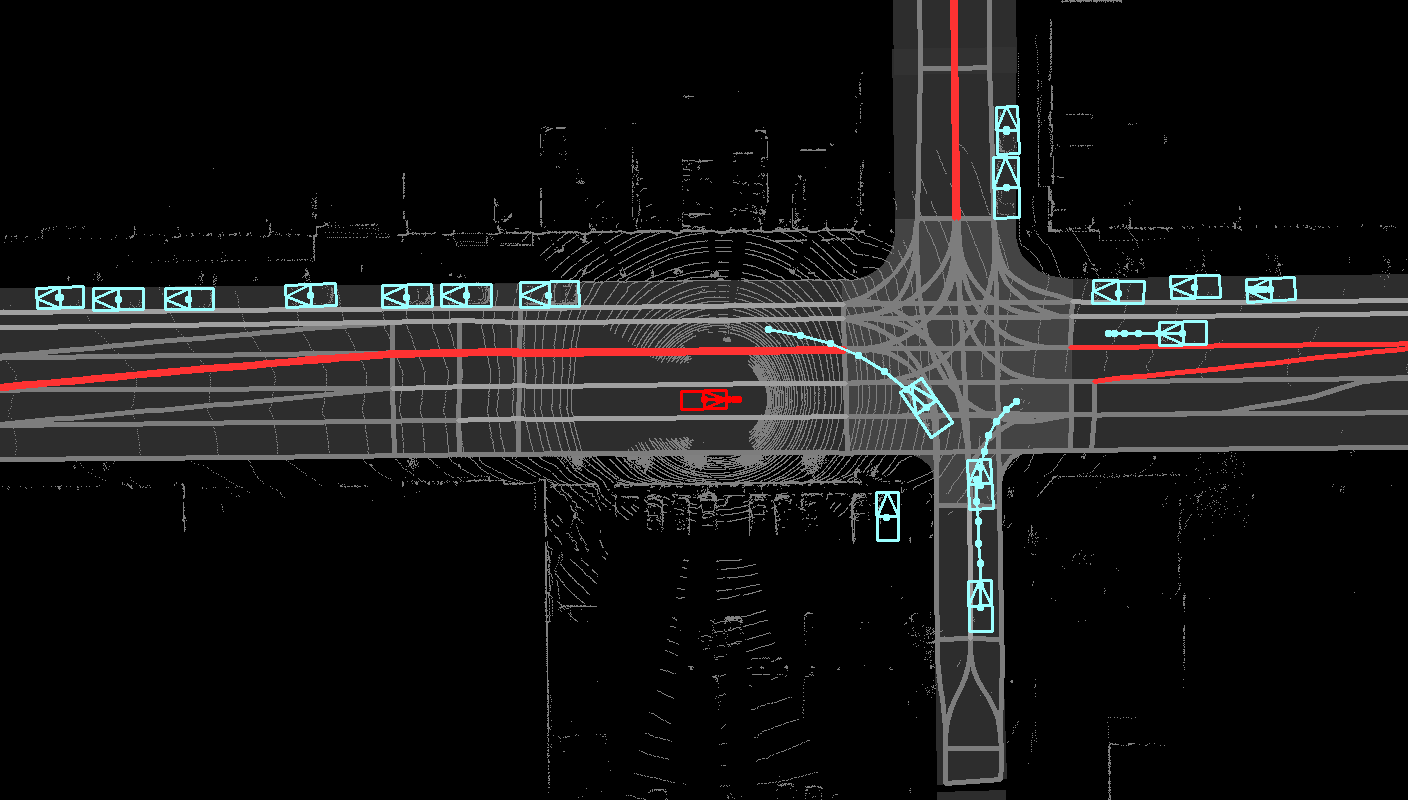}\\

\end{tabular}
\caption{Left: Input to our model. Middle: Detection outputs (shown in cyan). Right: Socially consistent prediction (shown in cyan) and safe motion
planning (shown in red).}
\label{fig:vis1}
\end{figure*}

\begin{figure*}[h]
\centering
\setlength{\tabcolsep}{1pt}
\begin{tabular}{ccc}
  \includegraphics[width=0.32\linewidth]{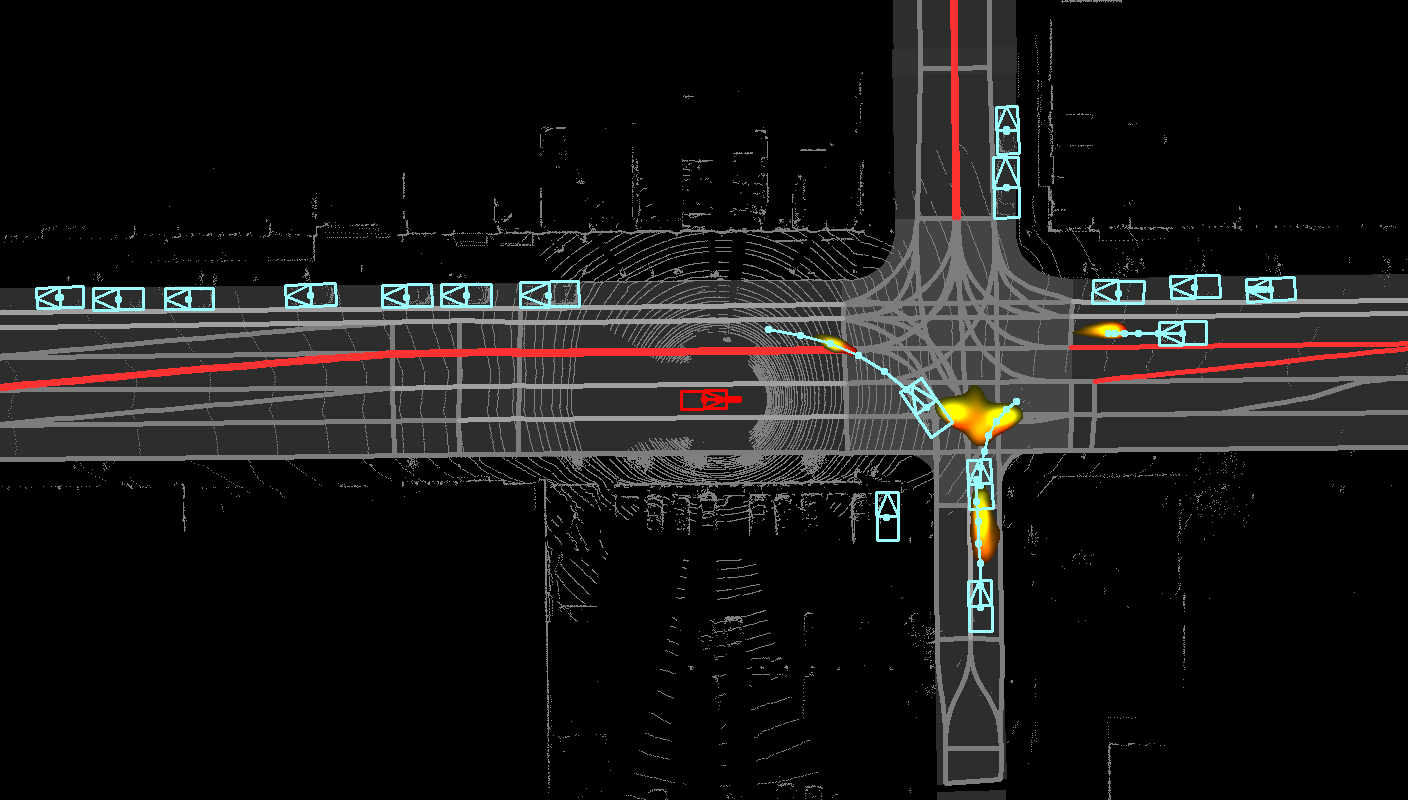}
  \includegraphics[width=0.32\linewidth]{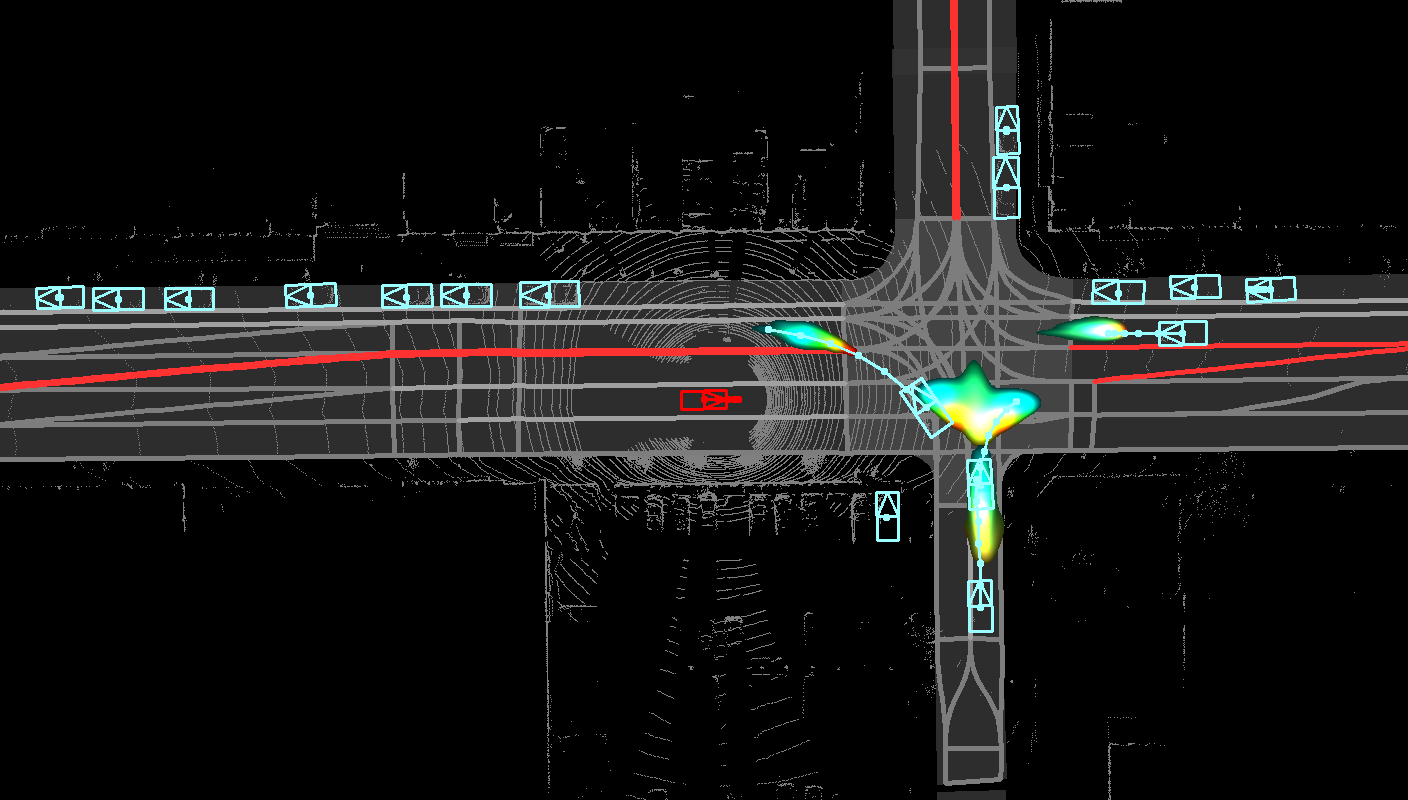}
  \includegraphics[width=0.32\linewidth]{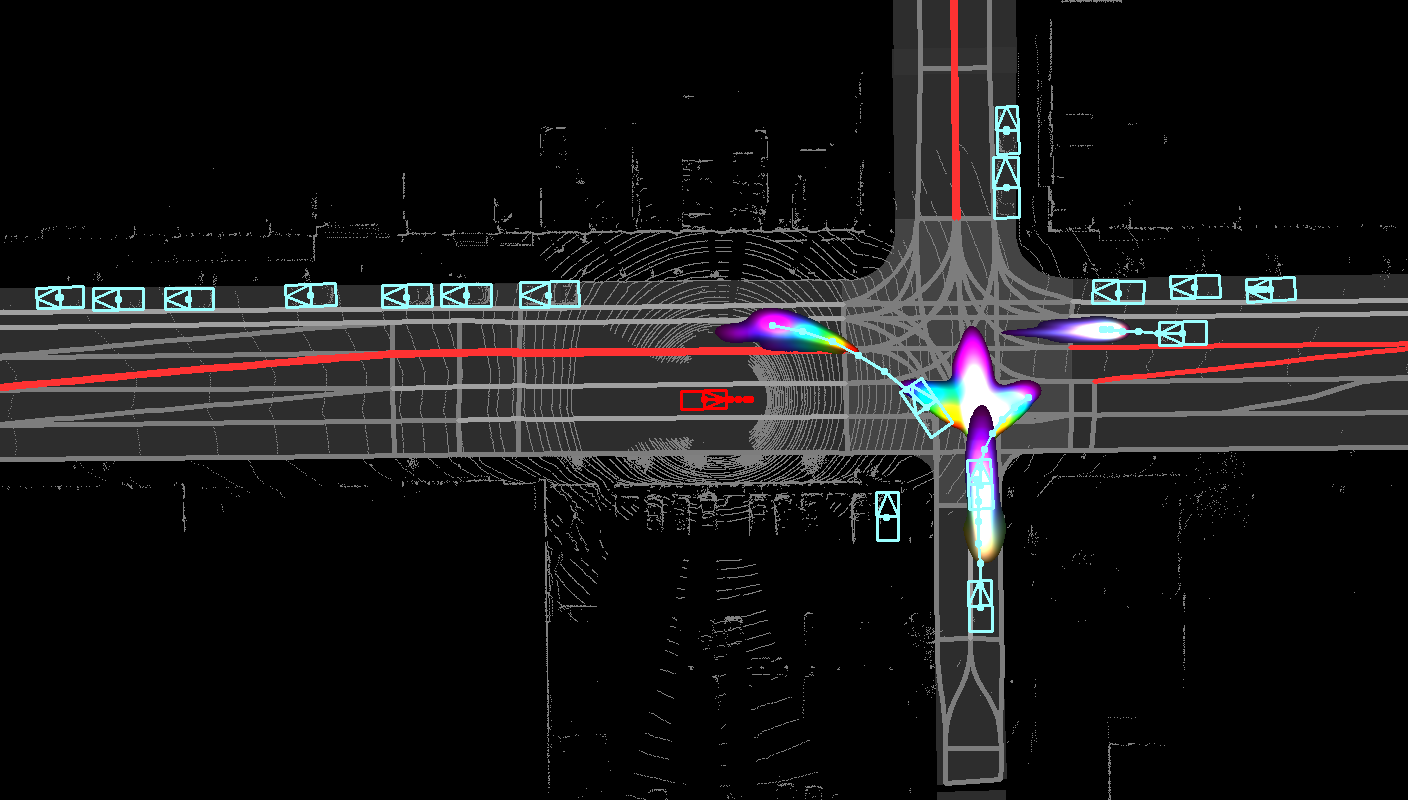}\\

\end{tabular}
\caption{We visualize the prediction uncertainty estimated by our model. We highlight the high-probability regions that actors might go to at
different future timestamps using different colors, and overlay them together to have a better visualization. From left to right: high-probability
region at 1 to 3 seconds into the future. We can observe clear multi-modality for the actor near an intersection (going straight / turning left / turning right).}
\label{fig:vis2}
\end{figure*}

\begin{figure*}[h]
\centering
\setlength{\tabcolsep}{1pt}
\begin{tabular}{cc}
  \includegraphics[width=0.48\linewidth]{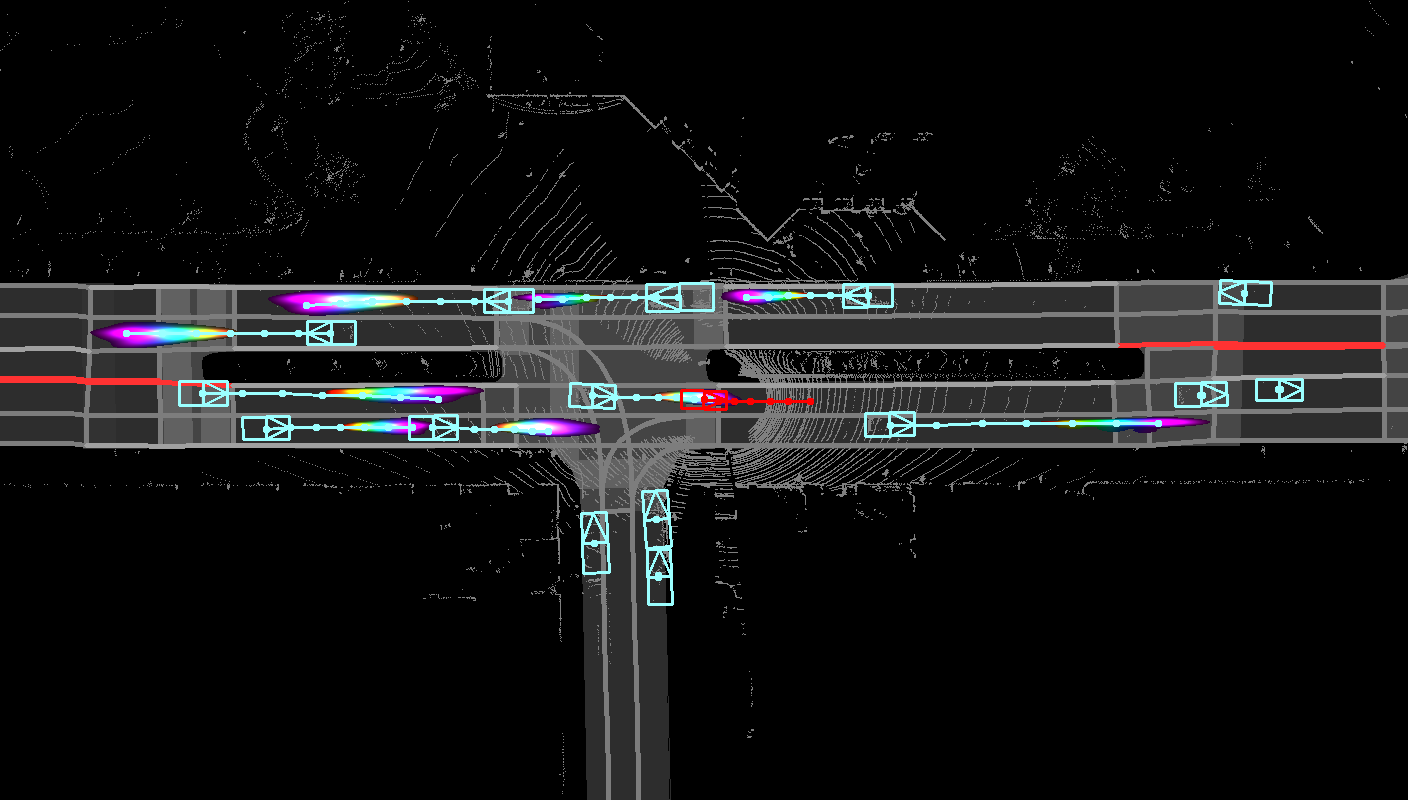}
  \includegraphics[width=0.48\linewidth]{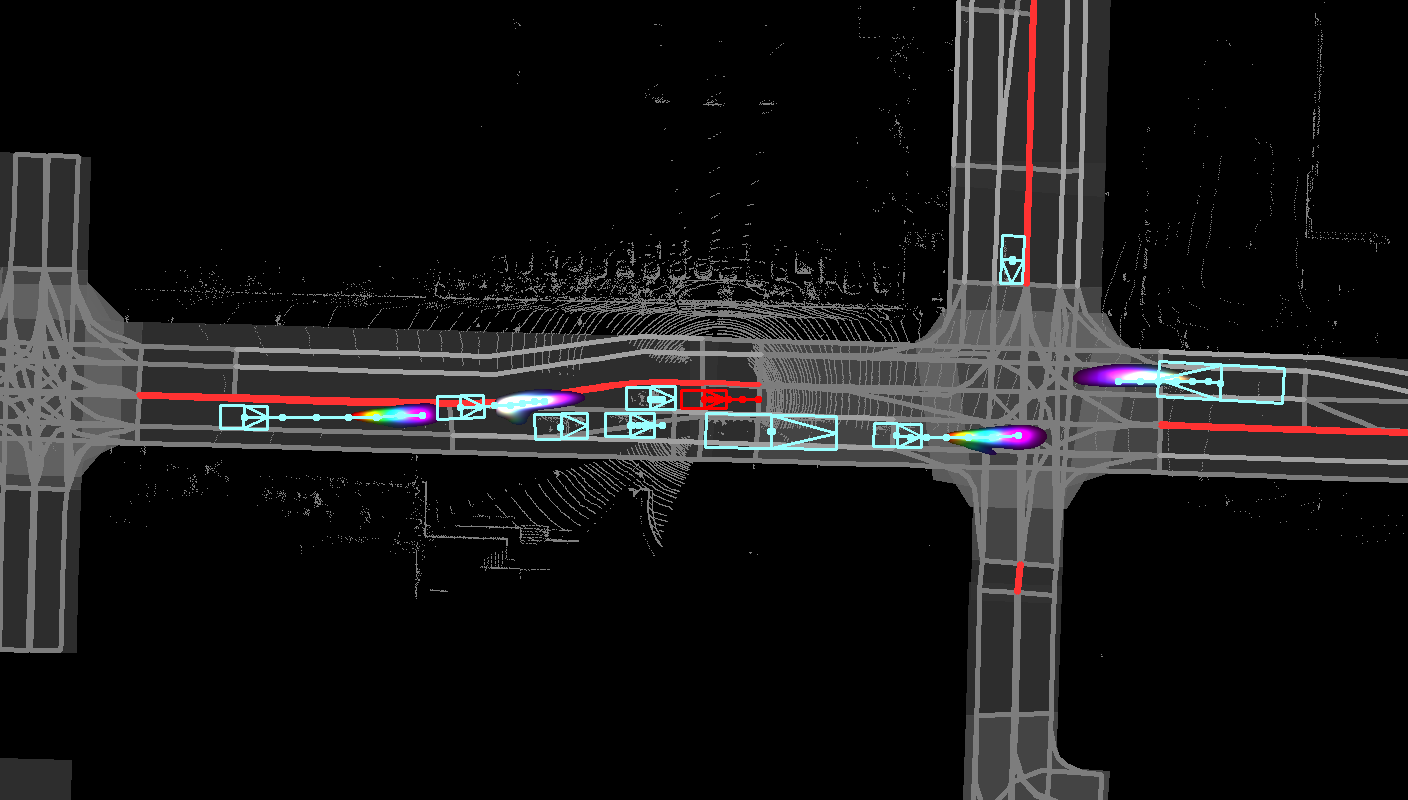}\\
  \includegraphics[width=0.48\linewidth]{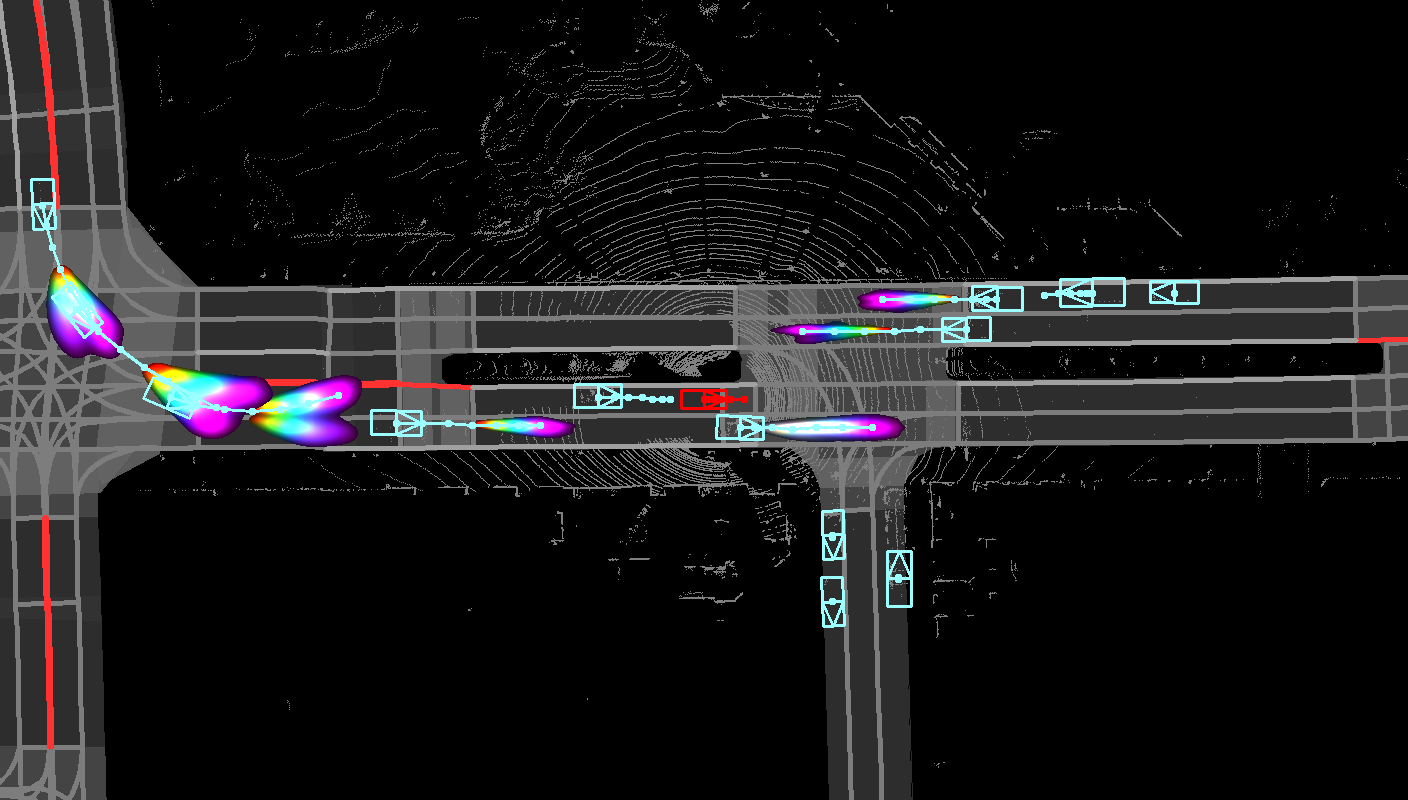}
  \includegraphics[width=0.48\linewidth]{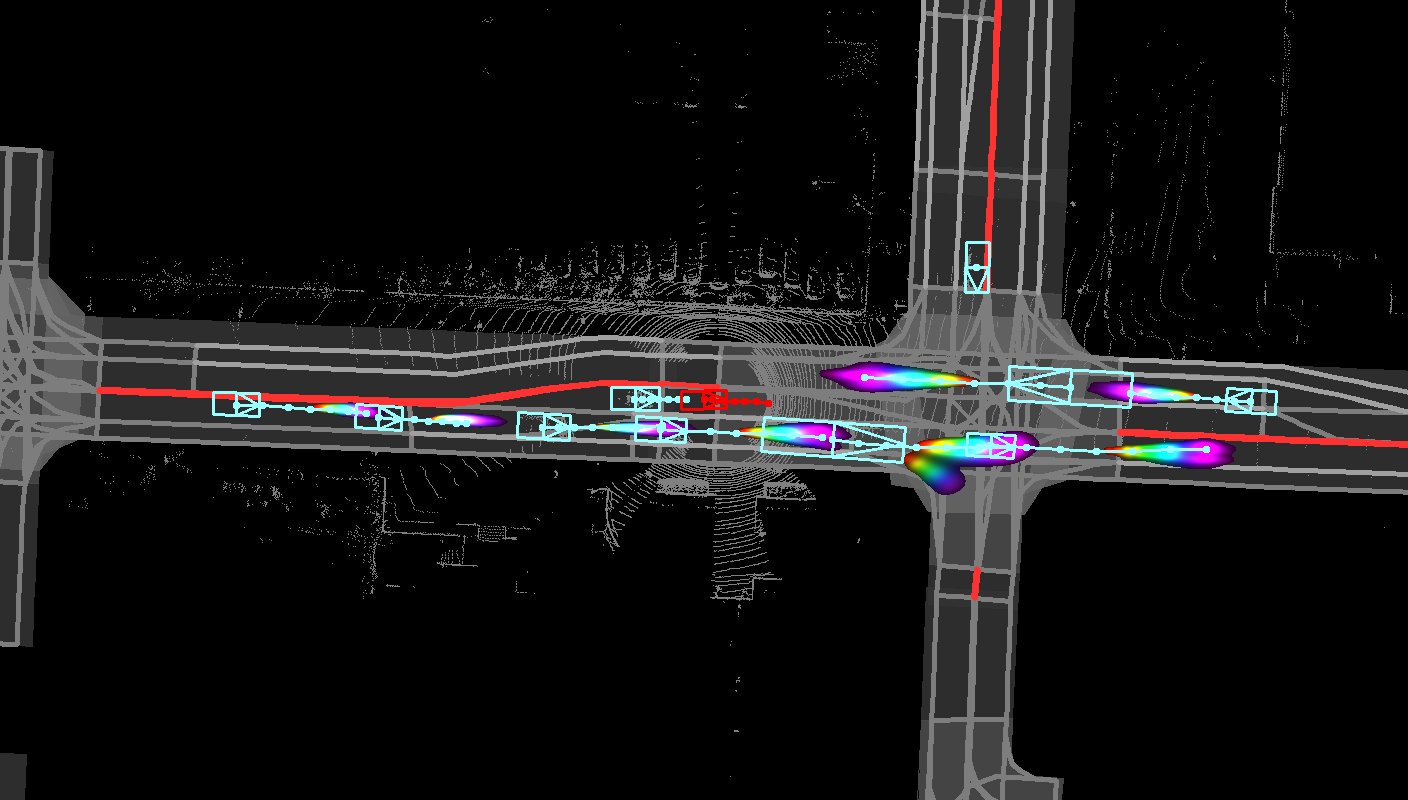}\\
  \includegraphics[width=0.48\linewidth]{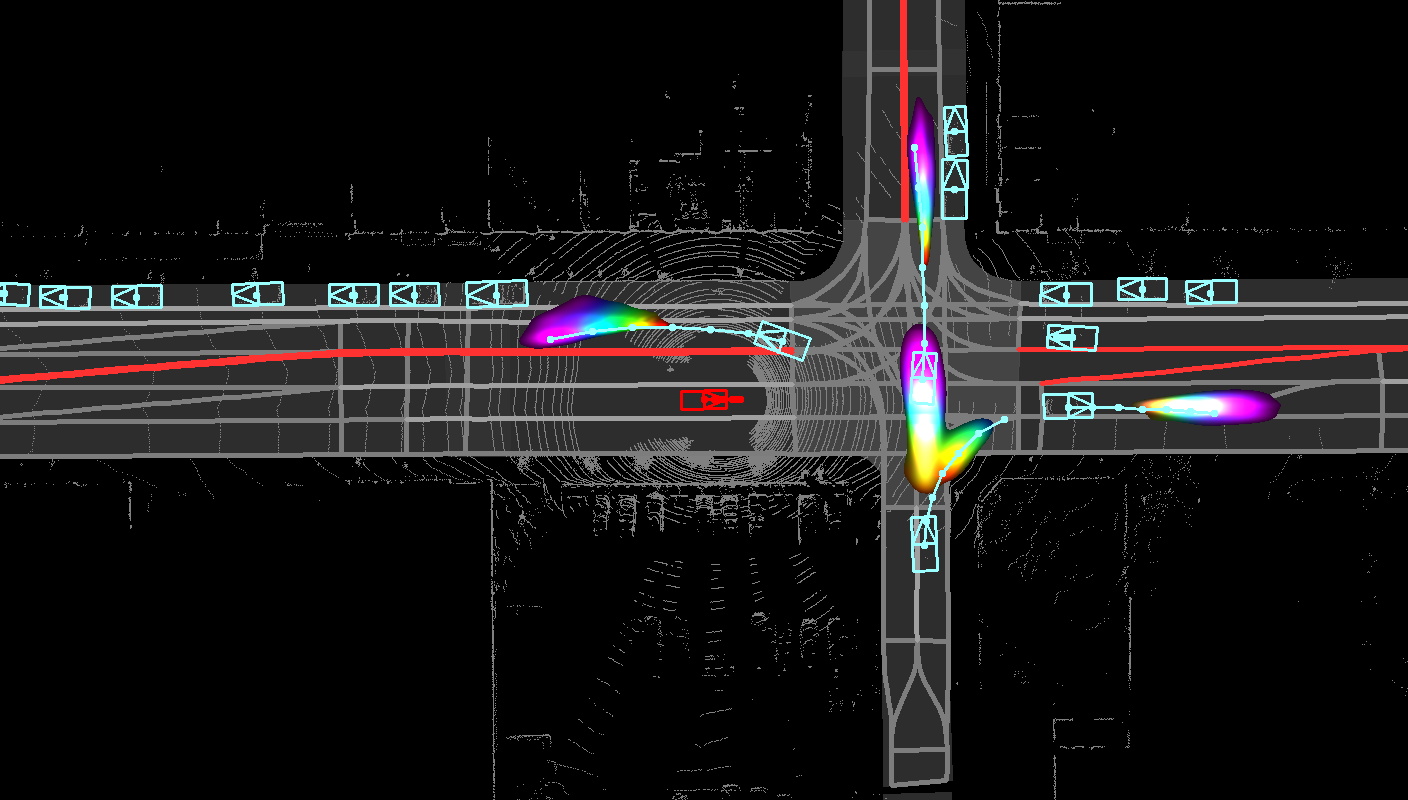}
  \includegraphics[width=0.48\linewidth]{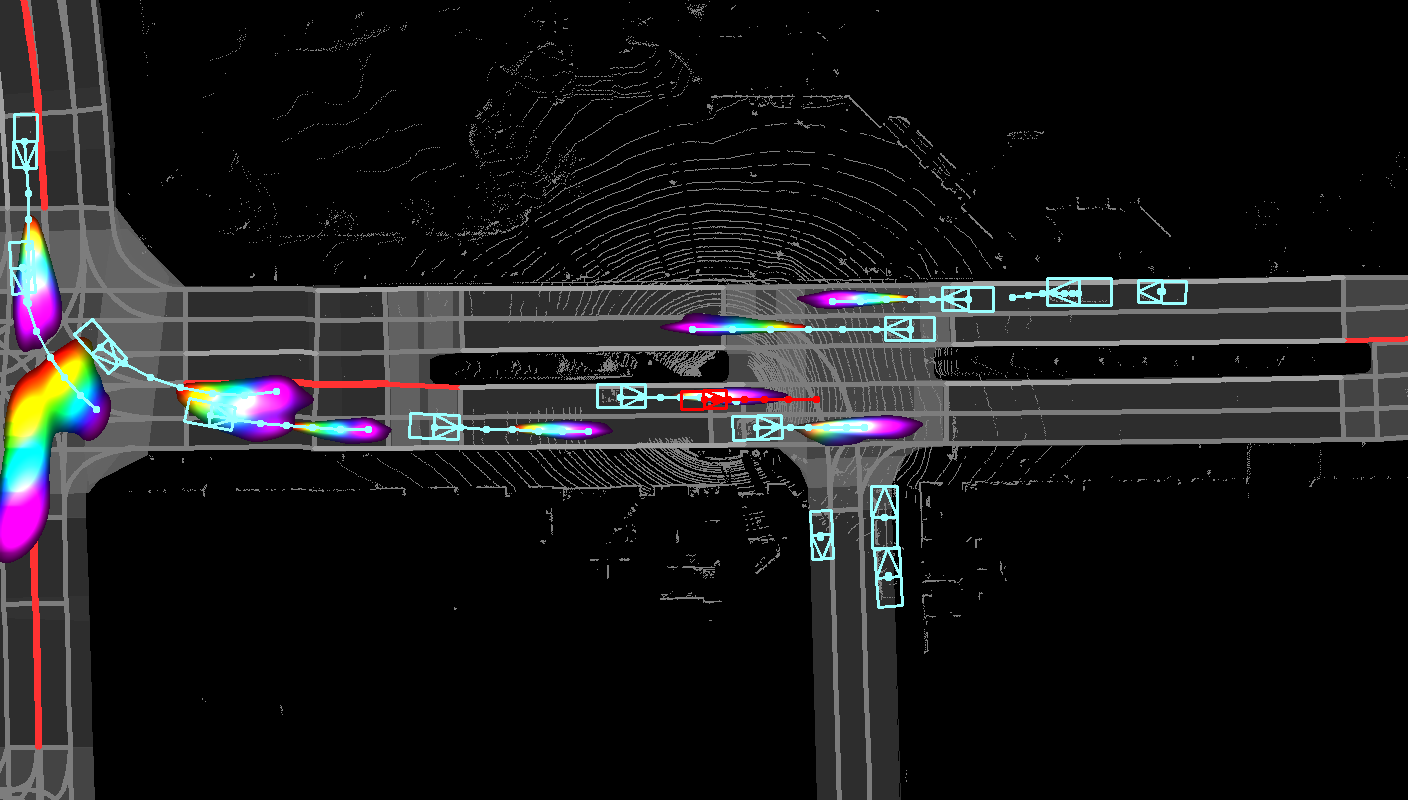}\\
  \includegraphics[width=0.48\linewidth]{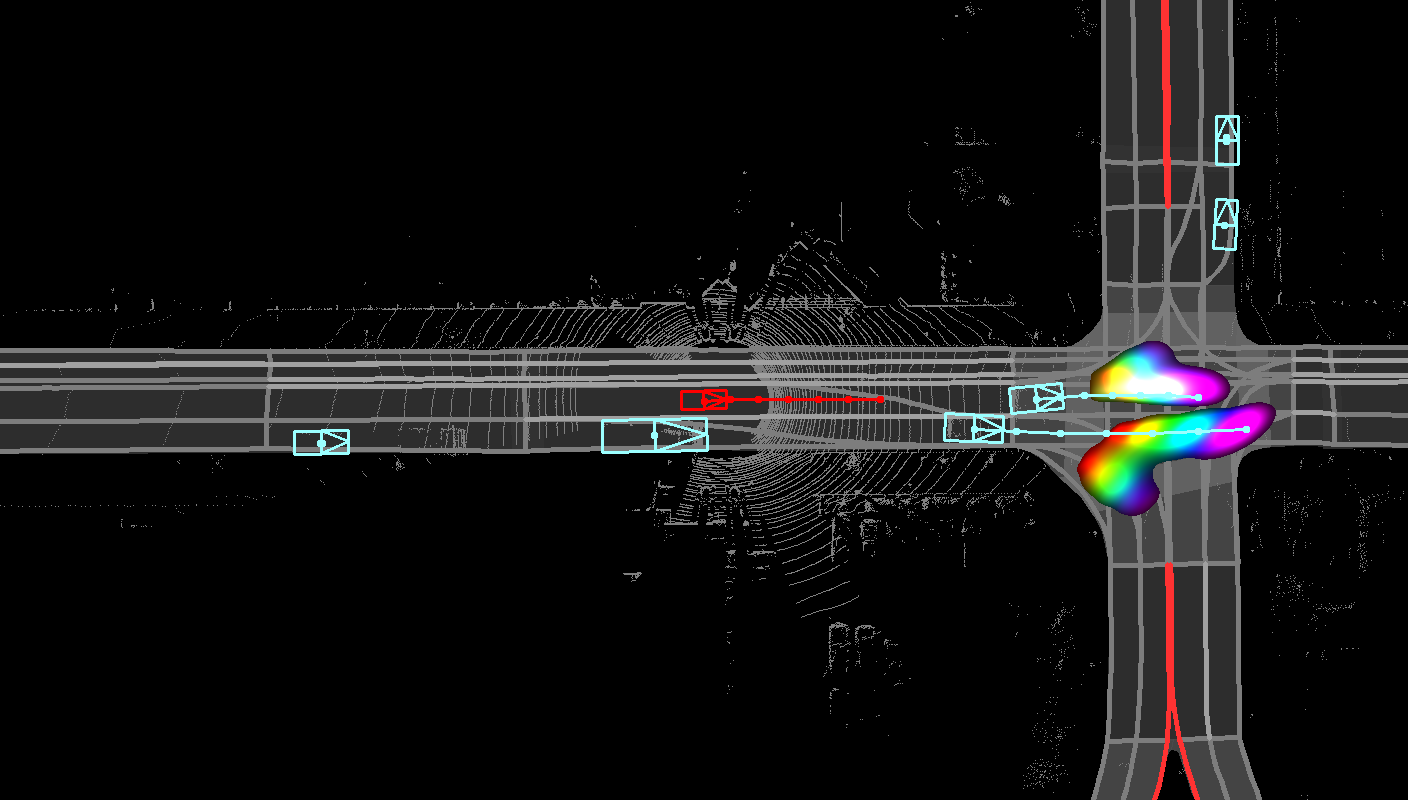}
  \includegraphics[width=0.48\linewidth]{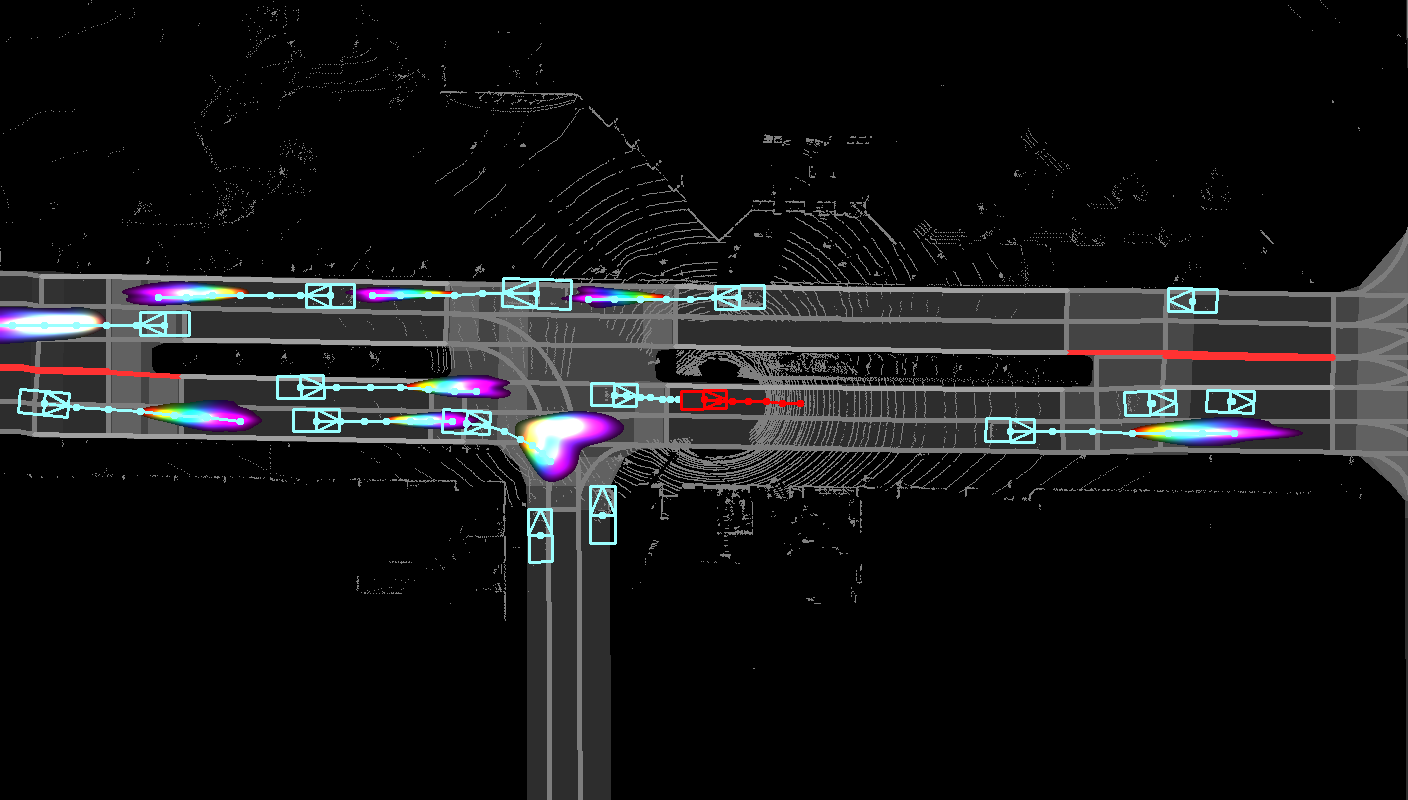}\\
\end{tabular}
\caption{More Prediction Visualizations. We show our predicted uncertainty: 1) aligns with lanes when an actor is driving on a straight road (meaning
  that we are certain about future direction but not certain about future speed in this case). 2) shows multi-modality when an actor approaches an intersection (either turning or going
straight).}
\label{fig:vis3}
\end{figure*}

\begin{figure*}[h]
\centering
\setlength{\tabcolsep}{1pt}
\begin{tabular}{cc}
  \includegraphics[width=0.48\linewidth]{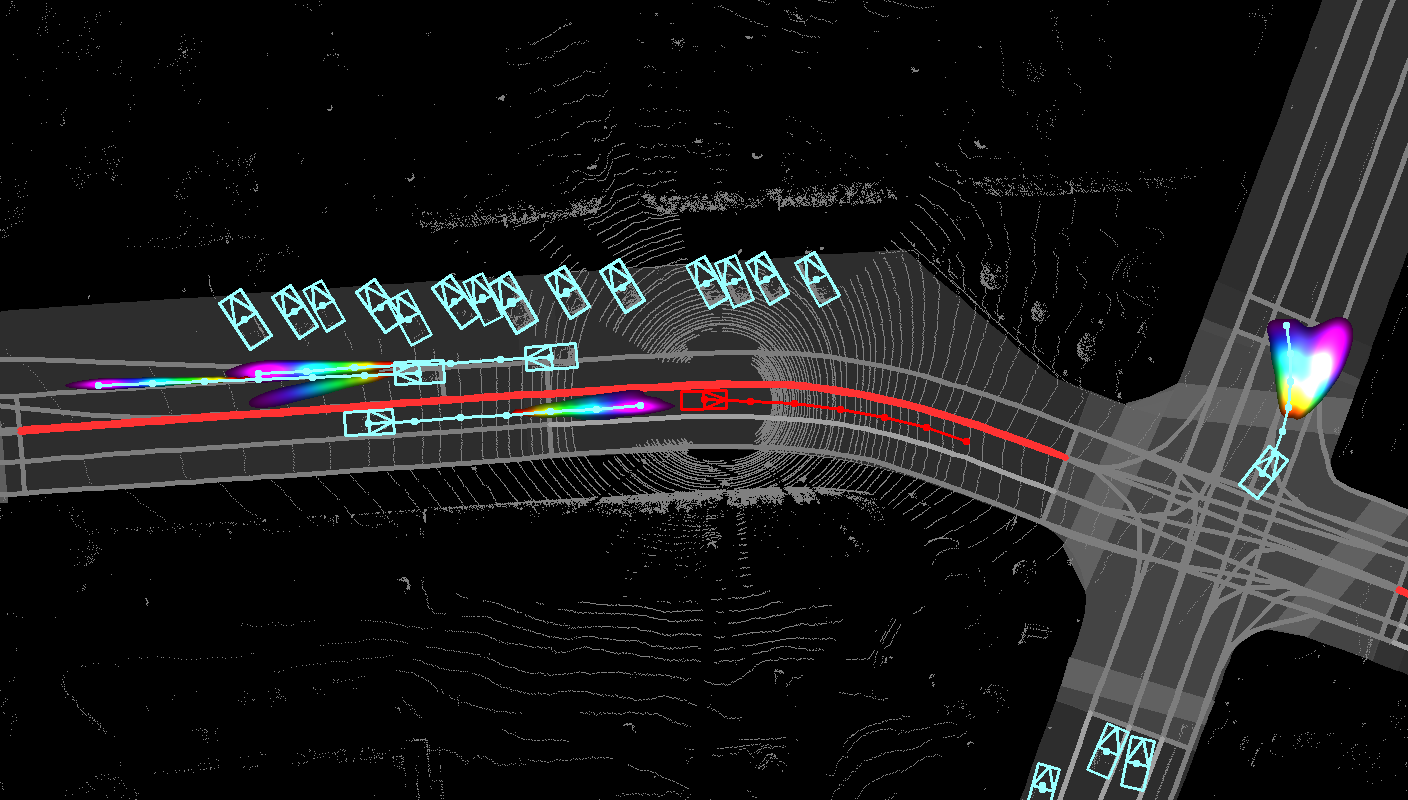}
  \includegraphics[width=0.48\linewidth]{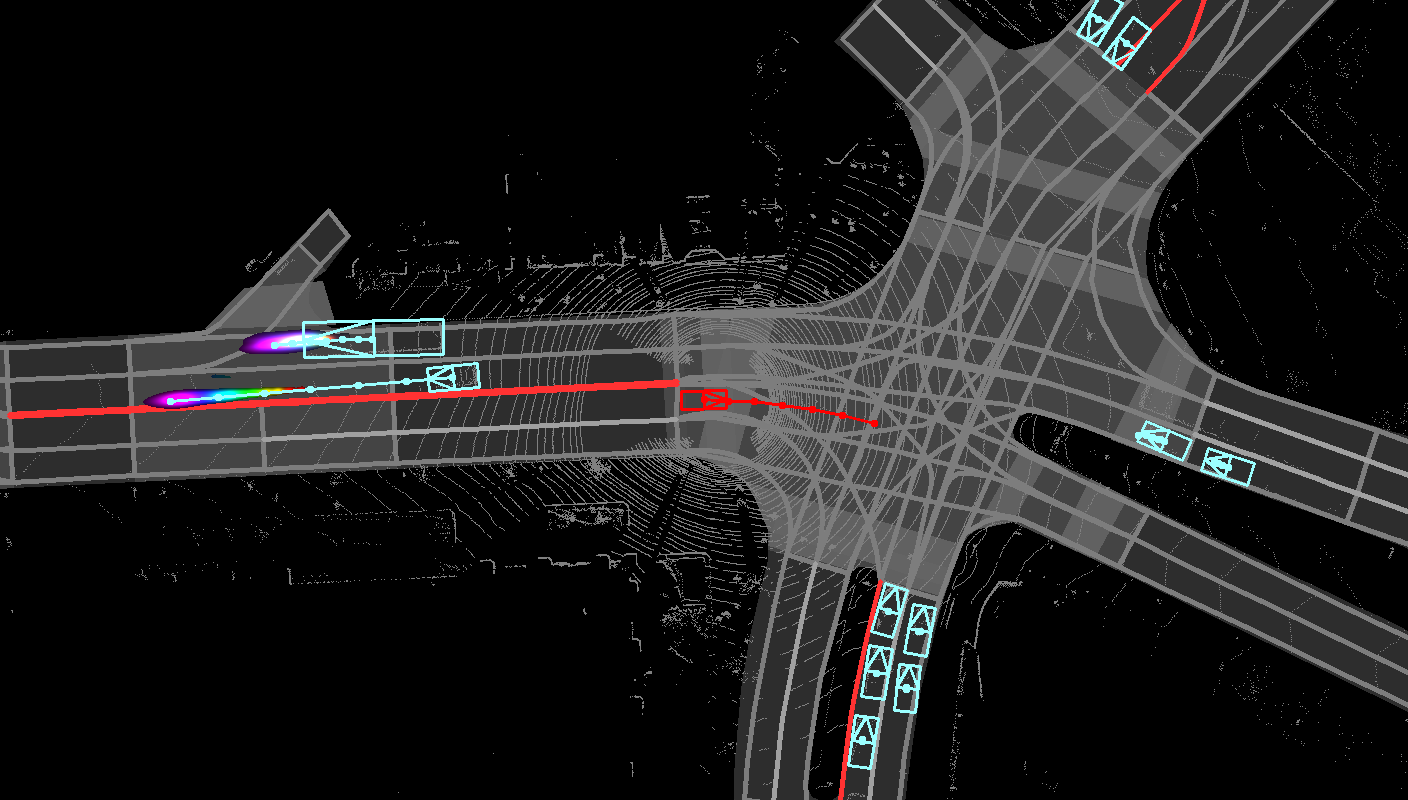}\\
  \includegraphics[width=0.48\linewidth]{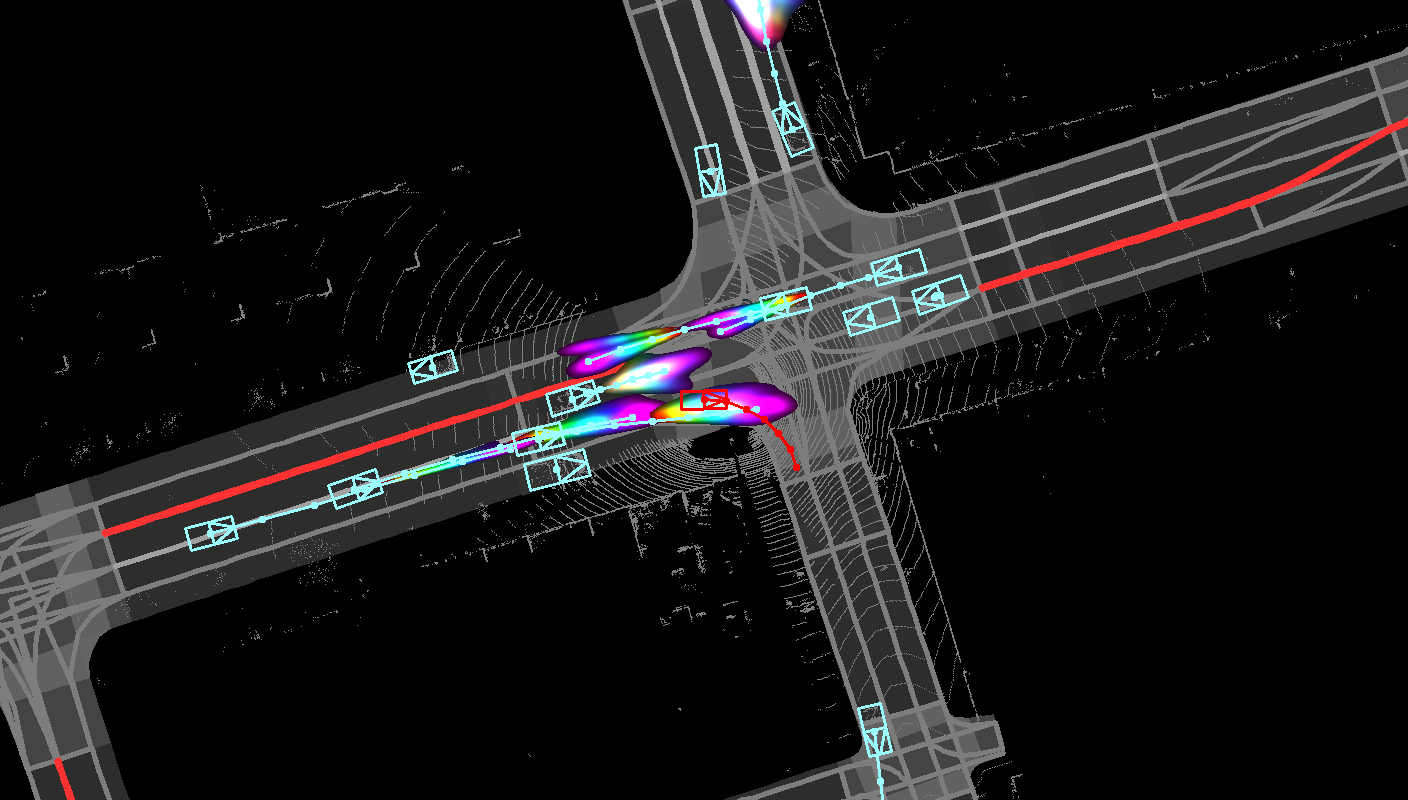}
  \includegraphics[width=0.48\linewidth]{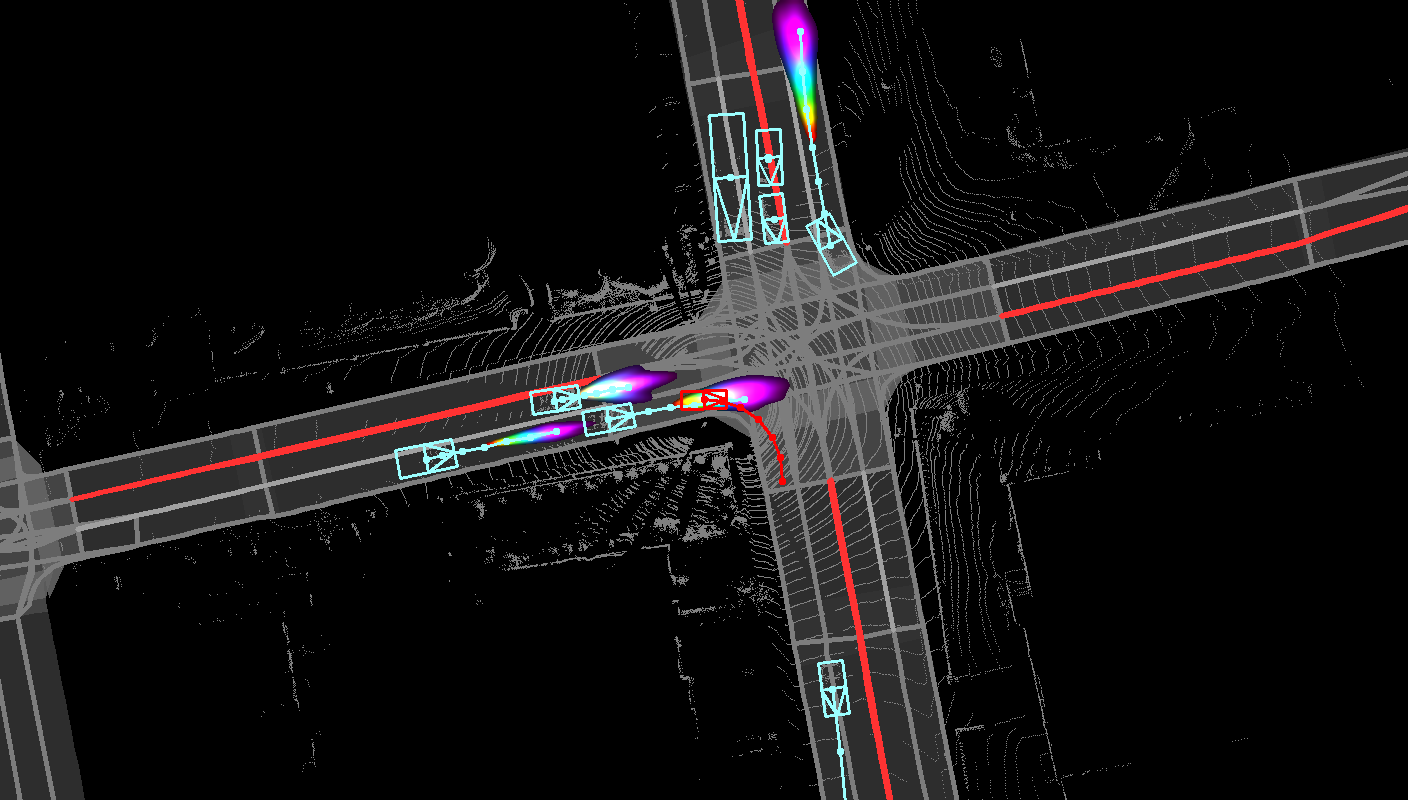}\\
  \includegraphics[width=0.48\linewidth]{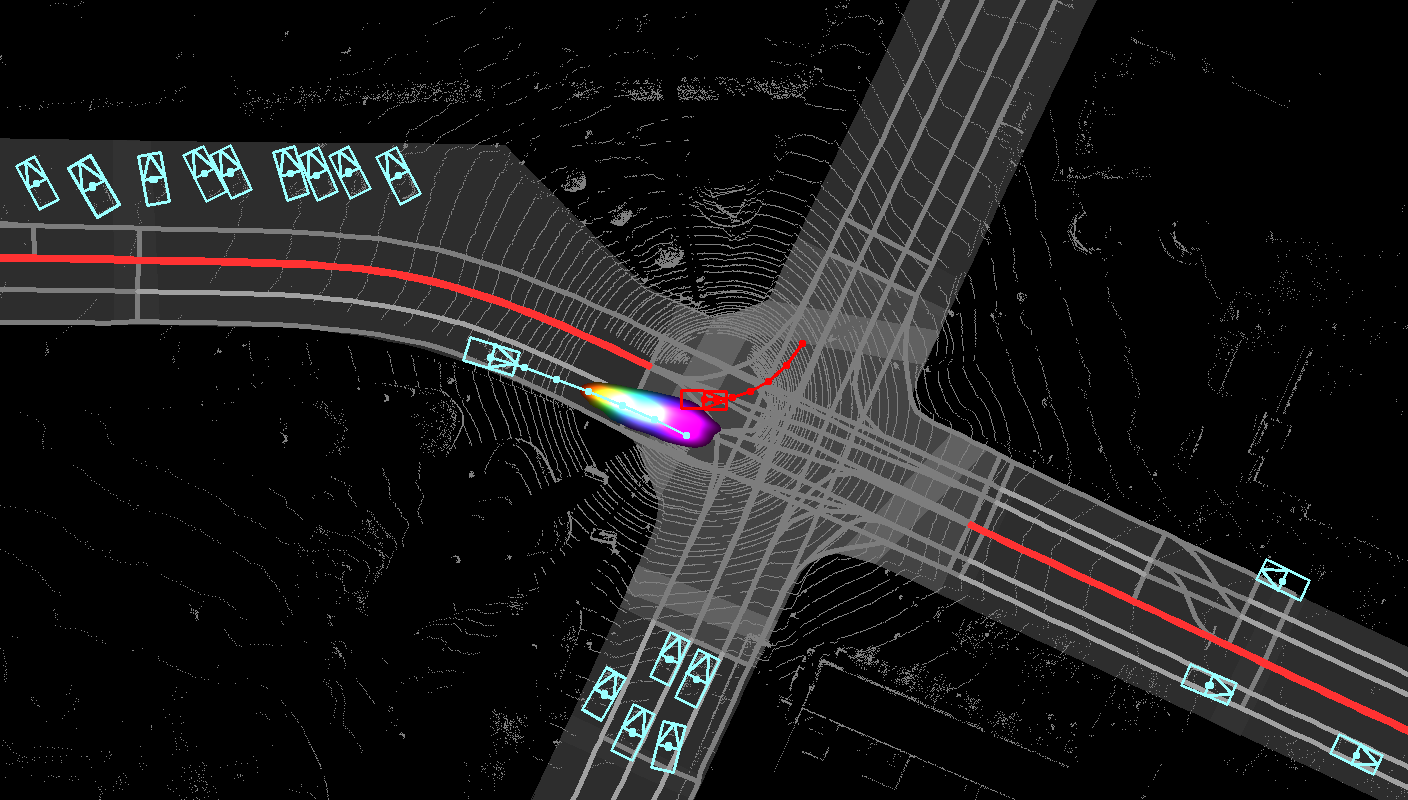}
  \includegraphics[width=0.48\linewidth]{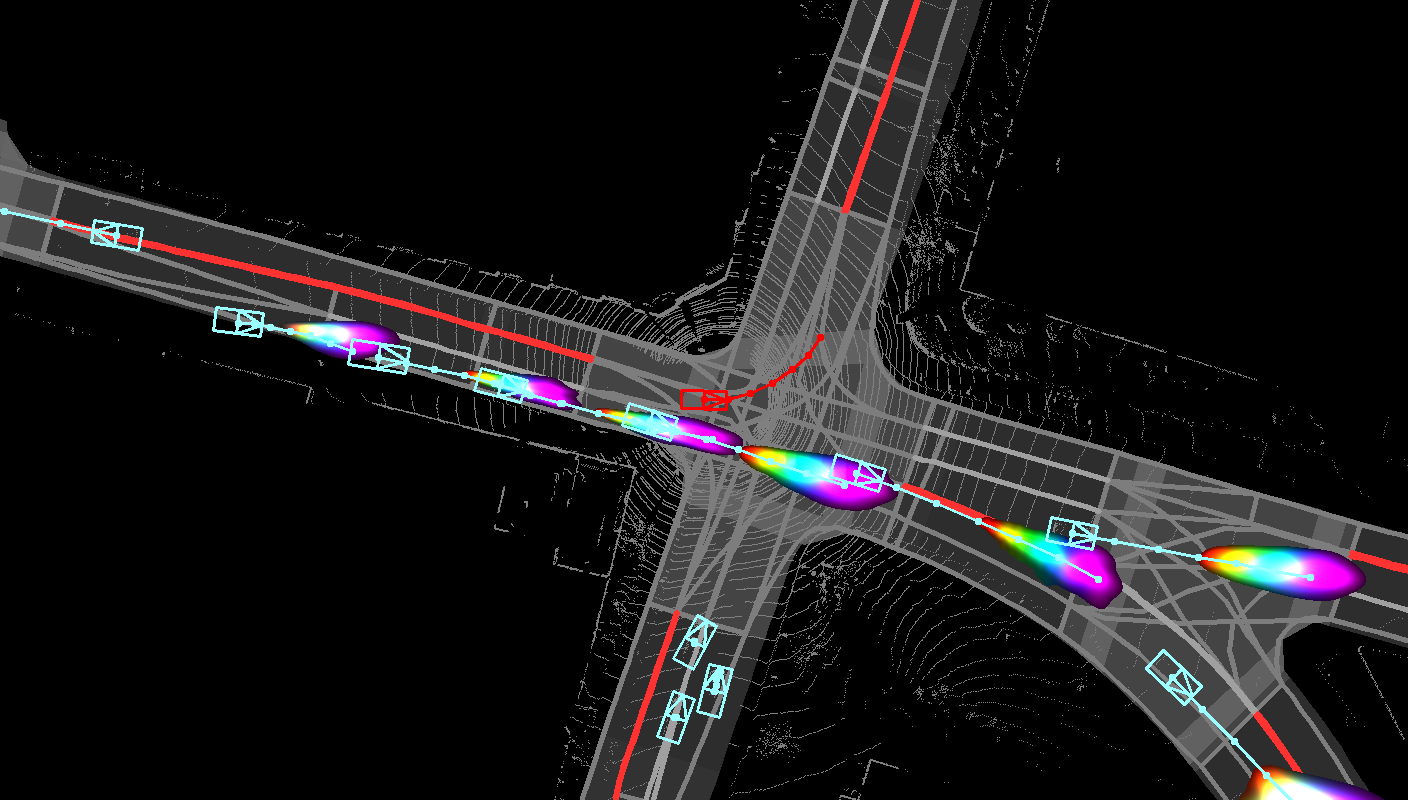}\\
  \includegraphics[width=0.48\linewidth]{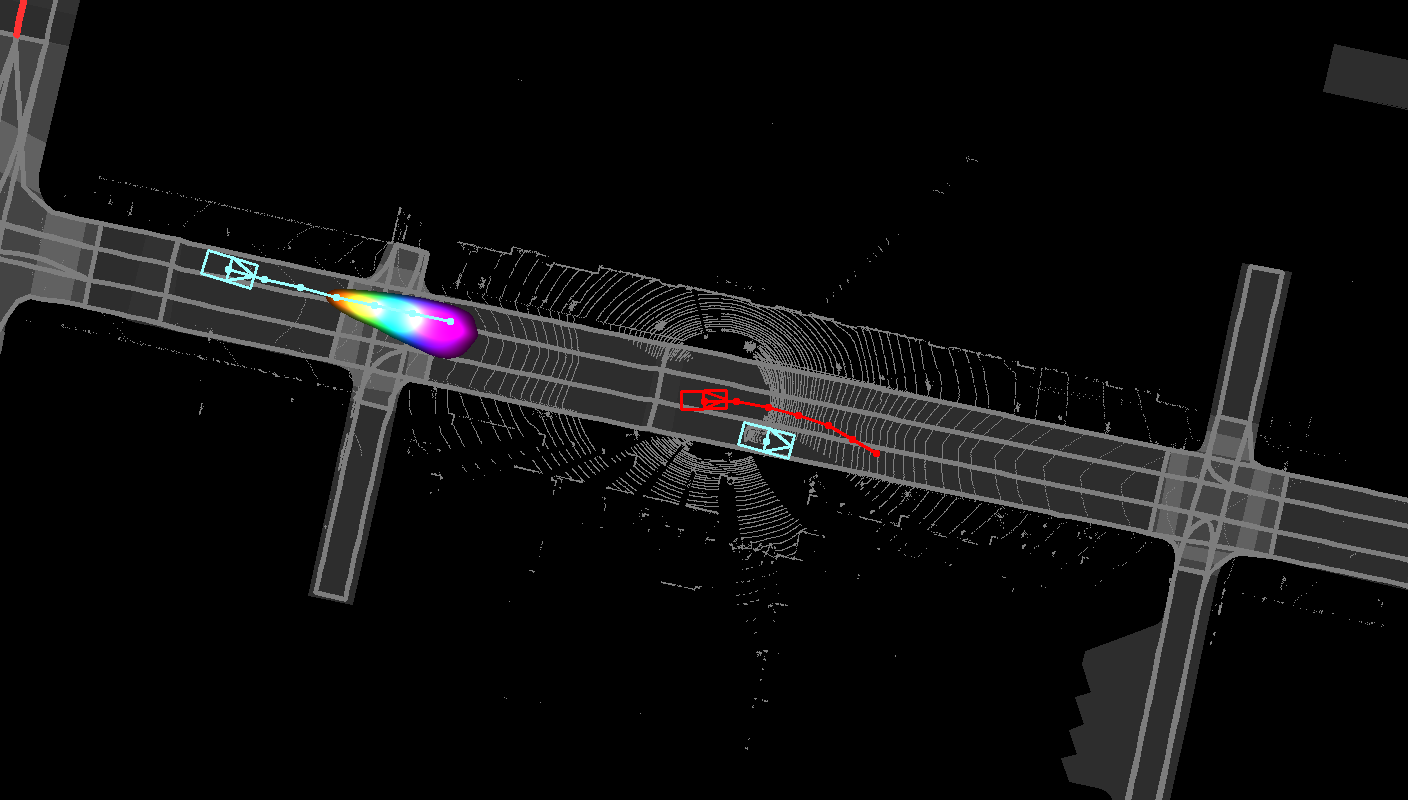}
  \includegraphics[width=0.48\linewidth]{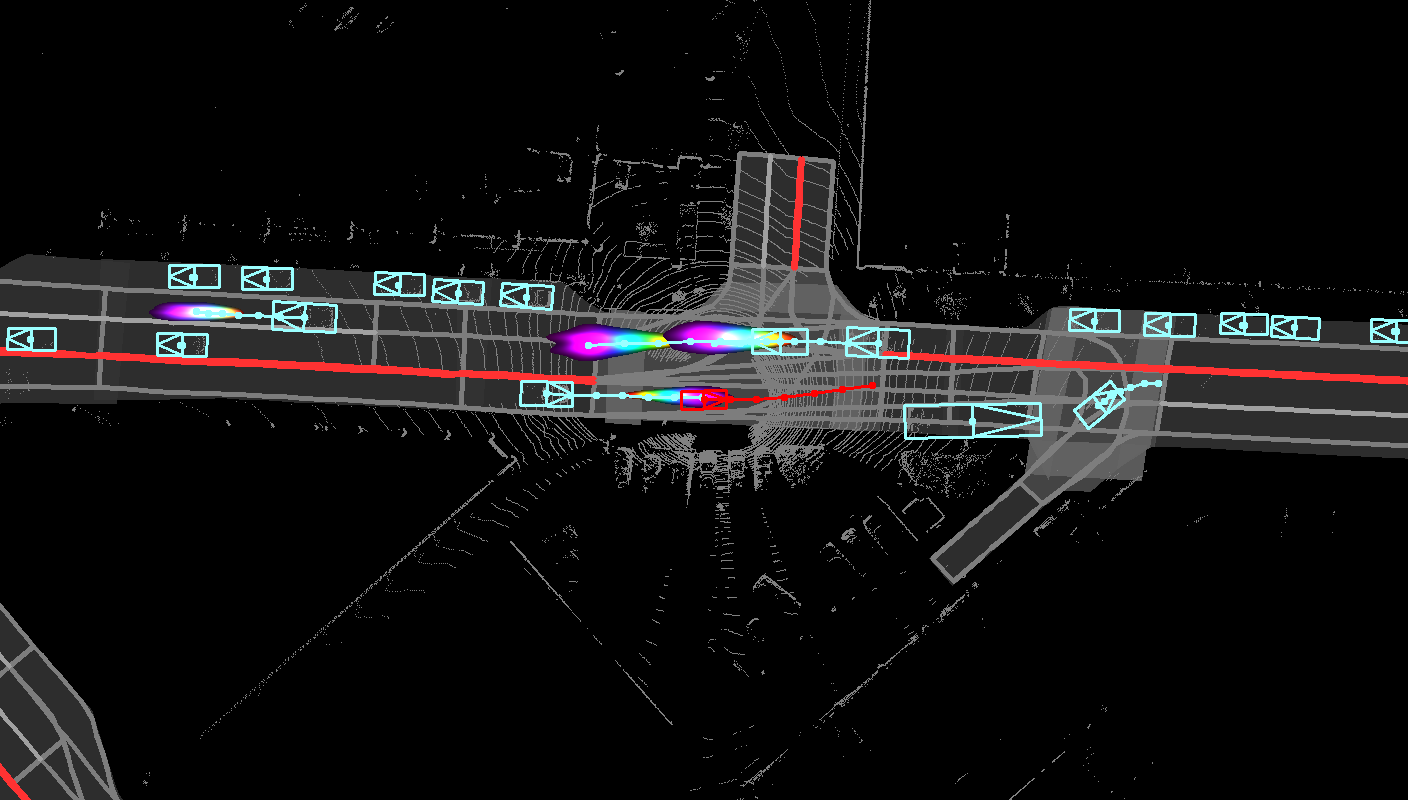}\\
\end{tabular}
\caption{More Planning Visualizations: We show our motion planner can nicely handle lane following (row 1), turning (row 2 \& 3) and nudging to avoid
collision (row 4).}
\label{fig:vis4}
\end{figure*}

\end{document}